\documentclass[]{article}

\usepackage{microtype}
\PassOptionsToPackage{numbers, compress}{natbib}

\usepackage[final]{neurips_2020}

\usepackage[utf8]{inputenc} \usepackage[T1]{fontenc}    \usepackage{url}            \usepackage{booktabs}       \usepackage{amsfonts}       \usepackage{nicefrac}       \usepackage{microtype}

\usepackage[utf8]{inputenc} \usepackage[T1]{fontenc}    
\usepackage[usenames,dvipsnames]{xcolor}
\usepackage[colorlinks,linktoc=all]{hyperref}
\usepackage[all]{hypcap}
\hypersetup{citecolor=MidnightBlue}
\hypersetup{linkcolor=MidnightBlue}
\hypersetup{urlcolor=MidnightBlue}

\usepackage{url}            \usepackage{booktabs}       \usepackage{amsfonts}       \usepackage{nicefrac}       \usepackage{microtype}      \usepackage{amsmath}
\usepackage{lipsum}
\usepackage{url}
\usepackage{subfigure}
\usepackage[subfigure]{tocloft}
\usepackage{pgfplots}
\usepackage{booktabs}
\usepackage{colortbl}
\usepackage[english]{babel} 
\usepackage[]{hyperref}
\usepackage{bm}             \usepackage{url}
\usepackage{placeins}
\usepackage[,ruled,noend]{algorithm2e}
\usepackage[font=small,labelfont=bf,tableposition=top]{caption}
\usepackage[bottom]{footmisc}
\usepackage{amsthm}
\usepackage{lscape}
\usepackage{xargs}          \usepackage{siunitx}
\usepackage{textcomp}
\usepackage{catchfile}
\usepackage{lipsum} 
\usepackage{float}
\usepackage{blindtext}
\usepackage[noend]{algpseudocode}
\usepackage{mathtools}

\DeclarePairedDelimiter{\ceil}{\lceil}{\rceil}

\newlength\figureheight
\newlength\figurewidth

\addto\extrasenglish{
}

\DeclareCaptionLabelFormat{andtable}{#1~#2  \&  \tablename~\thetable}
\usepgfplotslibrary{groupplots}
\usetikzlibrary{arrows,shapes,automata,backgrounds,petri,positioning}
\usetikzlibrary{decorations.pathmorphing}
\usetikzlibrary{decorations.shapes}
\usetikzlibrary{decorations.text}
\usetikzlibrary{decorations.fractals}
\usetikzlibrary{decorations.footprints}
\usetikzlibrary{shadows}
\usetikzlibrary{calc}
\usetikzlibrary{spy}
\tikzstyle{every picture}+=[font=\sffamily]
\tikzstyle{optimized} = [circle,fill=white,draw=black, dashed,inner sep=1pt,
minimum size=20pt, font=\fontsize{10}{10}\selectfont, node distance=1]
\pgfplotsset{
  tick label style = {font=\sffamily},
  every axis label/.append style={font=\sffamily},
}
\pgfplotsset{every axis/.append style={
      every y tick scale label/.style={at={(rel axis cs:0,1)},anchor=south,xshift=1ex},
      major tick length=1ex,
      minor tick length=1ex,
			every x tick label/.append style={font=\fontsize{5pt}{5pt}\sffamily, yshift=.5ex,},
			every y tick label/.append style={font=\fontsize{5pt}{5pt}\sffamily, xshift=.8ex},
			every y label/.append style={xshift=10ex}
		},
}

\pgfplotsset{compat=newest}
\usetikzlibrary{external}
\tikzexternaldisable

\usepackage[capitalize,nameinlink]{cleveref}
\crefname{section}{\S}{\S\S}
\Crefname{section}{\S}{\S\S}
\creflabelformat{equation}{#2\textup{#1}#3}

\definecolor{mygreen}{rgb}{0.2, 0.7, 0.2}
\definecolor{myorange}{rgb}{0.9, 0.5, 0.0}

\definecolor{blue}{HTML}{5DA5DA}
\definecolor{orange}{HTML}{FAA43A} 
\definecolor{green}{HTML}{60BD68} 
\definecolor{pink}{HTML}{F17CB0} 
\definecolor{brown}{HTML}{B2912F} 
\definecolor{purple}{HTML}{B276B2} 
\definecolor{yellow}{HTML}{DECF3F} 
\definecolor{red}{HTML}{F15854} 
\definecolor{gray}{HTML}{4D4D4D}

\def\N{{\mathcal{N}}}

 \newcommand{\R}{\mathbb{R}}

\newcommand{\T}{\top}

\newcommand{\diag}{\mathrm{diag}}

\newcommand{\Tr}{\mathrm{Tr}}

\newcommand{\norm}{\mathcal{N}}

\newcommand{\fvect}{\mathbf{f}}
\newcommand{\gvect}{\mathbf{g}}
\newcommand{\Ivect}{\mathbf{I}}

\newcommand{\hvect}{\mathbf{h}}

\newcommand{\mvect}{\mathbf{m}}

\newcommand{\svect}{\mathbf{s}}

\newcommand{\uvect}{\mathbf{u}}

\newcommand{\vvect}{\mathbf{v}}

\newcommand{\zvect}{\mathbf{z}}
\newcommand{\xvect}{\mathbf{x}}

\newcommand{\yvect}{\mathbf{y}}

\newcommand{\wvect}{\mathbf{w}}

\newcommand{\zerovect}{\mathbf{0}}

\newcommand{\Din}{{D_\mathrm{in}}}
\newcommand{\Dout}{{D_\mathrm{out}}}

\newcommand{\thetavect}{{\boldsymbol{\theta}}}

\newcommand{\phivect}{\boldsymbol{\phi}}
\newcommand{\Phivect}{\boldsymbol{\Phi}}

\newcommand{\lambdavect}{\boldsymbol{\lambda}}

\newcommand{\muvect}{{\boldsymbol{\mu}}}

\newcommand{\sigmavect}{\boldsymbol{\sigma}}
\def\Sigmavect{{\bm{\Sigma}}}

\newcommand{\varepsilonvect}{\bm{\epsilon}}

\def\vect{{\mathrm{vect}}}
\newcommand{\bigO}{\mathcal{O}}

\newcommand{\name}[1]{{\textsc{#1}}\xspace}

\newcommand{\vi}{\name{vi}}

\def\fastfood{\name{fastfood}}
\newcommand{\tensorflow}{\textsc{tensorflow}\xspace}
\newcommand{\pytorch}{\name{pytorch}}

\newcommand{\gp}{\name{gp}}
\newcommand{\gps}{\textsc{gp}s\xspace}

\newcommand{\dnn}{\name{dnn}}
\newcommand{\dnns}{\textsc{dnn}s\xspace}

\newcommand{\cnn}{\name{cnn}}

\newcommand{\bnn}{\name{bnn}}

\newcommand{\cnns}{\textsc{cnn}s\xspace}
\newcommand{\mcd}{\name{mcd}}

\newcommand{\relu}{{\textsc{r}}e\name{lu}}

\newcommand{\rbf}{\name{rbf}}

\newcommand{\mnll}{\name{mnll}}
\newcommand{\ece}{\name{ece}}
\newcommand{\errate}{\name{error rate}}

\newcommand{\rmse}{\name{rmse}}
\newcommand{\nelbo}{\name{nelbo}}
\newcommand{\elbo}{\name{elbo}}

\newcommand{\nf}{\name{nf}}

\newcommand{\cifart}{\name{cifar10}}

\def\hvi{\name{whvi}}

\newcommand{\resnet}{\name{resnet}}

\newcommand{\alexnet}{\name{AlexNet}}
\newcommand{\vgg}{\name{vgg16}}

\newcommand{\Exp}{\mathbb{E}}

\newcommand{\KL}{\name{kl}}

\DeclareMathAlphabet{\mathsfit}{\encodingdefault}{\sfdefault}{m}{sl}
\newcommand{\tens}[1]{\mathbf{{#1}}}

\def\Wtens{{\tens{W}}}

\def\Amatr{{\bm{A}}}
\def\Bmatr{{\bm{B}}}

\def\Gmatr{{\bm{G}}}
\def\Hmatr{{\bm{H}}}
\def\Imatr{{\bm{I}}}

\def\Kmatr{{\bm{K}}}
\def\Lmatr{{\bm{L}}}
\def\Mmatr{{\bm{M}}}

\def\Qmatr{{\bm{Q}}}

\def\Smatr{{\bm{S}}}
\def\Tmatr{{\bm{T}}}
\def\Umatr{{\bm{U}}}
\def\Vmatr{{\bm{V}}}
\def\Wmatr{{\bm{W}}}
\def\Xmatr{{\bm{X}}}
\def\Ymatr{{\bm{Y}}}

\def\Phimatr{{\bm{\Phi}}}
\def\Pimatr{{\bm{\Pi}}}

\def\Gammamatr{{\bm{\Gamma}}}
\def\Sigmamatr{{\bm{\Sigma}}}
\def\Omegamatr{{\bm{\Omega}}}

\begin{document}
 
\makeatletter
\newcommand{\loadtikz}[1]{
    \filename@parse{#1}
    \let\fpath\filename@area
    \ifdefined\compilefigures
        \newcommand{\sometext}{\input{\filename@area\filename@base}}
        \tikzsetnextfilename{\filename@area\filename@base.pdf}
        \sometext
    \else
        \includegraphics[]{001_\fpath\filename@baseXpdfXpdf}
    \fi
}

\makeatother
\title{Walsh-Hadamard Variational Inference\\for Bayesian Deep Learning}

\author{Simone Rossi\thanks{Equal contribution} \\
  Data Science Department\\
  EURECOM (FR)\\
  \texttt{simone.rossi@eurecom.fr} \\
  \And
  S\'ebastien Marmin\textsuperscript{*} \\
  Data Science Department\\
  EURECOM (FR)\\
  \texttt{sebastien.marmin@eurecom.fr} \\
  \And
  Maurizio Filippone \\
  Data Science Department\\
  EURECOM (FR)\\
  \texttt{maurizio.filippone@eurecom.fr} \\
}
\maketitle

\begin{abstract}
    Over-parameterized models, such as DeepNets and ConvNets, form a class of models that are routinely adopted in a wide variety of applications, and for which Bayesian inference is desirable but extremely challenging.
Variational inference offers the tools to tackle this challenge in a scalable way and with some degree of flexibility on the approximation, but for over-parameterized models this is challenging due to the over-regularization property of the variational objective.  
Inspired by the literature on kernel methods, and in particular on structured approximations of distributions of random matrices, this paper proposes Walsh-Hadamard Variational Inference (\hvi), which uses Walsh-Hadamard-based factorization strategies to reduce the parameterization and accelerate computations, thus avoiding over-regularization issues with the variational objective.
Extensive theoretical and empirical analyses demonstrate that \hvi yields considerable speedups and model reductions compared to other techniques to carry out approximate inference for over-parameterized models, and ultimately show how advances in kernel methods can be translated into advances in approximate Bayesian inference for Deep Learning.

\end{abstract}

\section{Introduction}
\label{sec:introduction}
Since its inception, Variational Inference (\vi, \citep{Jordan99}) has continuously gained popularity as a scalable and flexible approximate inference scheme for a variety of models for which exact Bayesian inference is intractable.
Bayesian neural networks \citep{Mackay94,Neal1997} represent a good example of models for which inference is intractable, and for which \vi -- and approximate inference in general -- is challenging due to the nontrivial form of the posterior distribution and the large dimensionality of the parameter space \citep{Graves11,Gal16}. 
Recent advances in \vi allow one to effectively deal with these issues in various ways. 
For instance, a flexible class of posterior approximations can be constructed using, e.g., normalizing flows \citep{Rezende2015}, whereas the need to operate with large parameter spaces has pushed the research in the direction of Bayesian compression \citep{Louizos2017a,Molchanov2017}.

Employing \vi is notoriously challenging for over-parameterized statistical models. 
In this paper, we focus in particular on Bayesian Deep Neural Networks (\dnns) and Bayesian Convolutional Neural Networks (\cnns) as typical examples of over-parameterized models.
Let's consider a supervised learning task with $N$ input vectors and corresponding labels collected in $\Xmatr = \{\xvect_1, \ldots, \xvect_N \}$ and $\Ymatr = \{\yvect_1, \ldots, \yvect_N \}$, respectively; furthermore, let's consider \dnns with weight matrices $\Wtens = \left\{\Wmatr^{(1)}, \ldots, \Wmatr^{(L)} \right\}$, likelihood $p(\Ymatr | \Xmatr, \Wtens)$, and prior $p(\Wtens)$.
Following standard variational arguments, after introducing an approximation $q(\Wtens)$ to the posterior $p(\Wtens | \Xmatr, \Ymatr)$ it is possible to obtain a lower bound to the log-marginal likelihood $\log\left[p(\Ymatr | \Xmatr)\right]$ as follows:
\begin{align}
\label{eq:lowerbound}
\log\left[p(\Ymatr | \Xmatr)\right] \geq \Exp_{q(\Wtens)}[ \log p(\Ymatr | \Xmatr, \Wtens)  ]
- \KL\{q(\Wtens) \| p(\Wtens)\}\,.
\end{align}
The first term acts as a model fitting term, whereas the second one acts as a regularizer, penalizing solutions where the posterior is far away from the prior.
It is easy to verify that the \KL term can be the dominant one in the objective for over-parameterized models. 
For example, a mean field posterior approximation turns the \KL term into a sum of as many \KL terms as the number of model parameters, say $Q$, which can dominate the overall objective when $Q \gg N$.  
As a result, the optimization focuses on keeping the approximate posterior close to the prior, disregarding the rather important model fitting term. 
This issue has been observed in a variety of deep models \citep{Bowman2016}, where it was proposed to gradually include the \KL term throughout the optimization \citep{Bowman2016, Sonderby2016} to scale up the model fitting term \citep{Wilson2020, Wenzel2020} or to improve the initialization of variational parameters \citep{Rossi2018}.
Alternatively, other approximate inference methods for deep models with connections to \vi have been proposed, notably Monte Carlo Dropout \citep[\mcd;][]{Gal16} and Noisy Natural Gradients \citep[\name{nng};][]{Zhang2018}. 

In this paper, we propose a novel strategy to cope with model over-parameterization when using variational inference, which is inspired by the literature on kernel methods. 
Our proposal is to reparameterize the variational posterior over model parameters by means of a structured decomposition based on random matrix theory \citep{Tropp2011}, which has inspired a number of fundamental contributions in the literature on approximations for kernel methods, such as \fastfood \citep{Le13} and Orthogonal Random Features (\name{orf}, \citep{Yu16}). 
The key operation within our proposal is the Walsh-Hadamard transform, and this is why we name our proposal Walsh-Hadamard Variational Inference (\hvi). 

Without loss of generality, consider Bayesian \dnns with weight matrices $\Wmatr^{(l)}$ of size $D \times D$.
 Compared with mean field \vi, \hvi has a number of attractive properties.
The number of parameters is reduced from $\bigO(D^2)$ to $\bigO(D)$, thus reducing the over-regularization effect of the \KL term in the variational objective.
We derive expressions for the reparameterization and the local reparameterization tricks, showing that, the computational complexity is reduced from $\bigO(D^2)$ to $\bigO(D \log{D})$.
Finally, unlike mean field \vi, \hvi induces a matrix-variate distribution to approximate the posterior over the weights, thus increasing flexibility at a log-linear cost in $D$ instead of linear.

We can think of our proposal as a specific factorization of the weight matrix, so we can speculate that other tensor factorizations \citep{Novikov2015} of the weight matrix could equally yield such benefits.
Our comparison against various matrix factorization alternatives, however, shows that \hvi is superior to other parameterizations that have the same complexity.
Furthermore, while matrix-variate posterior approximations have been proposed in the literature of \vi \citep{Louizos2016}, this comes at the expense of increasing the complexity, while our proposal keeps the complexity to log-linear in $D$.

Through a wide range of experiments on \dnns and \cnns, we demonstrate that our approach enables the possibility to run variational inference on complex over-parameterized models, while being competitive with state-of-the-art alternatives.
Ultimately, our proposal shows how advances in kernel methods can be instrumental in improving \vi, much like previous works showed how kernel methods can improve, e.g., Markov chain Monte Carlo sampling \citep{Sejdinovic2014,Strathmann2015} and statistical testing \citep{Gretton2007,Gretton2012,Zaremba2013}.

\section{Walsh-Hadamard Variational Inference}
\label{sec:whvi}
\subsection{Background on Structured Approximations of Kernel Matrices}

\hvi is inspired by a line of works that developed from random feature expansions for kernel machines \citep{Rahimi08}, which we briefly review here.
A positive-definite kernel function $\kappa(\xvect_i, \xvect_j)$ induces a mapping $\phivect(\xvect)$, which can be infinite dimensional depending on the choice of $\kappa(\cdot,\cdot)$.
Among the large literature of scalable kernel machines, random feature expansion techniques aim at constructing a finite approximation to $\phivect(\cdot)$.
For many kernel functions \citep{Rahimi08, Cho09}, this approximation is built by applying a nonlinear transformation to a random projection $\Xmatr\Omegamatr$, where $\Omegamatr$ has entries $\N(\omega_{ij}|0,1)$.
If the matrix of training points $\Xmatr$ is $N \times D$ and we are aiming to construct $D$ random features, that is $\Omegamatr$ is $D \times D$, this requires $N$ times $\bigO(D^2)$ time, which can be prohibitive when $D$ is large.

\fastfood \citep{Le13} tackles the issue of large dimensional problems by replacing the matrix $\Omegamatr$ with a random matrix for which the space complexity is reduced from $\bigO(D^2)$ to $\bigO(D)$ and time complexity of performing products with input vectors is reduced from $\bigO(D^2)$ to $\bigO(D \log D)$.
In \fastfood, the matrix $\Omegamatr$ is replaced by 
$ \Omegamatr \approx \Smatr\Hmatr\Gmatr\Pimatr \Hmatr\Bmatr \text{,}
    \label{eq:fastfood}
$ where $\Pimatr$ is a permutation matrix, $\Hmatr$ is the Walsh-Hadamard matrix, whereas $\Gmatr$ and $\Bmatr$ are diagonal random matrices with standard Normal and Rademacher ($\{\pm 1\}$) distributions, respectively.
The Walsh-Hadamard matrix is defined recursively starting from
$\tiny
    H_2 = \begin{bmatrix}
        1 & 1  \\
        1 & -1 \\
    \end{bmatrix}
$ and then
$\tiny
    H_{2D} = \begin{bmatrix}
        H_D & H_D  \\
        H_D & -H_D \\
    \end{bmatrix}
$, possibly scaled by $D^{-1/2}$ to make it orthonormal.
The product $\Hmatr\xvect$ can be computed in $\bigO(D\log D)$ time and $\bigO(1)$ space using the in-place version of the Fast Walsh-Hadamard Transform \citep[\textsc{fwht},][]{Fino1976}.
$\Smatr$ is also diagonal with i.i.d. entries, and it is chosen such that 
the elements of $\Omegamatr$ obtained by this series of operations are approximately independent and follow a standard Normal (see \citep{Tropp2011} for more details).
\fastfood inspired a series of other works on kernel approximations
, whereby Gaussian random matrices are approximated by a series of products between diagonal Rademacher and Walsh-Hadamard matrices \citep{Yu16,Bojarski2017}.

\setlength\figureheight{.3\textwidth}
\setlength\figurewidth{.3\textwidth}
\begin{figure}[]
    \begin{minipage}[t]{.5\textwidth}
        \vspace{0ex}
        \centering
        \tiny
        \pgfplotsset{every axis title/.append style={yshift=-1ex}}
        \pgfplotsset{every x tick label/.append style={font=\fontsize{2}{4}\selectfont}}
        \pgfplotsset{every y tick label/.append style={font=\fontsize{2}{4}\selectfont}}
        \includegraphics{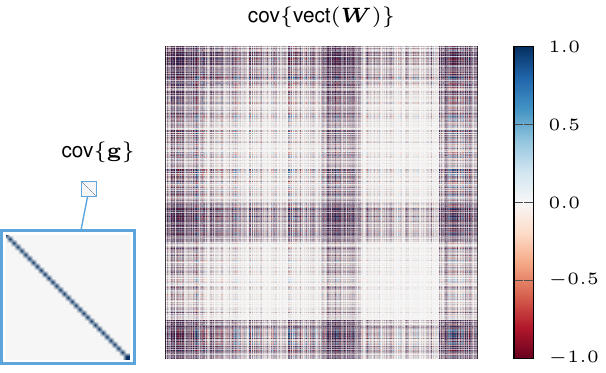}
        \captionof{figure}{Normalized covariance of $\gvect$ and $\vect(\Wmatr)$.}
        \label{fig:covariance}
    \end{minipage}\hfill\begin{minipage}[t]{.48\textwidth}
        \tiny
        \vspace{0ex}
        \captionof{table}{Complexity of various approaches to \vi
            \label{tab:complexity}
        }
        \sc
        \begin{center}
            \begin{tabular}{lcc}
                \toprule
                                        & \multicolumn{2}{c}{Complexity}                      \\
                {}                      & Space                          & Time               \\
                \midrule
                Mean field Gaussian     & $\bigO(D^2)$                   & $\bigO(D^2)$       \\
                Gaussian matrix variate & $\bigO(D^2)$                   & $\bigO(D^2 + M^3)$ \\
                Tensor factorization    & $\bigO(KR^2)$                  & $\bigO(R^2)$       \\
                \hvi                    & $\bigO(D)$                     & $\bigO(D\log D)$   \\
                \bottomrule
            \end{tabular}
        \end{center}
        \label{tab:complexity}
        \vspace{1ex}\par
        \scriptsize\normalfont
        Note: $D$ is the dimensionality of the feature map, $K$ is the number of tensor cores, $R$ is the rank of tensor cores and $M$ is the number of pseudo-data used to sample from a matrix Gaussian distribution (see \citep{Louizos2016}).
    \end{minipage}
    \end{figure}

\subsection{From \fastfood to Walsh-Hadamard Variational Inference}

\fastfood and its variants yield cheap approximations to Gaussian random matrices with pseudo-independent entries, and zero mean and unit variance.
The question we address in this paper is whether we can use these types of approximations as cheap approximating distributions for \vi.
By considering a prior for the elements of the diagonal matrix $\Gmatr = \diag(\gvect)$ and a variational posterior $q(\gvect) = \N(\muvect, \Sigmamatr)$, we can actually obtain a class of approximate posterior with some desirable properties as discussed next.
Let $\Wmatr = \Wmatr^{(l)} \in \R^{D \times D}$ be the weight matrix of a \dnn at layer $(l)$, and consider
\begin{align}
    \widetilde{\Wmatr} \sim q(\Wmatr) \quad \mathrm{s.t.} \quad \widetilde{\Wmatr} = \Smatr_1 \Hmatr \diag(\tilde\gvect) \Hmatr\Smatr_2 \quad
    \mathrm{with} \quad \widetilde{\gvect} \sim q(\gvect).
\end{align}
The choice of a Gaussian $q(\gvect)$ and the linearity of the operations induce a parameterization of a matrix-variate Gaussian distribution for $\Wmatr$, which is controlled by $\Smatr_1$ and $\Smatr_2$ if we assume that we can optimize these diagonal matrices.
Note that we have dropped the permutation matrix $\Pimatr$ and we will show later that this is not critical for performance, while it speeds up computations.

For a generic $D_1 \times D_2$ matrix-variate Gaussian distribution, we have
\begin{align}
    \Wmatr \sim \mathcal{MN}(\Mmatr, \Umatr, \Vmatr)
    \quad
    \text{if and only if}
    \quad
    \vect(\Wmatr) \sim \N(\vect(\Mmatr), \Vmatr \otimes \Umatr),
\end{align}
where $\Mmatr\in\R^{D_1 \times D_2}$ is the mean matrix and $\Umatr\in\R^{D_1 \times D_1}$ and $\Vmatr\in\R^{D_2 \times D_2}$ are two positive definite covariance matrices among rows and columns, and $\otimes$ denotes the Kronecker product.
In \hvi, as $\Smatr_2$ is diagonal, $\Hmatr\Smatr_2 = [\vvect_1, \dots, \vvect_{D}]$ with $\vvect_i = (\Smatr_2)_{i,i}(\Hmatr)_{:,i}$, so $\Wmatr$ can be rewritten in terms of $\Amatr\in\R^{D^2\times D}$ and $\gvect$ as follows
\begin{align} \label{eq:definition:Amatr}
    \vect(\Wmatr) = \Amatr\gvect \quad \mathrm{where} \quad
    \Amatr^{\T} =
    \left[
        (\Smatr_1\Hmatr\diag(\vvect_1))^{\T}
        \ldots
        (\Smatr_1\Hmatr\diag(\vvect_D))^{\T}
        \right].
\end{align}
This rewriting, shows that the choice of $q(\gvect)$ yields $q(\vect(\Wmatr)) = \N(\Amatr\muvect, \Amatr\Sigmamatr\Amatr^\T)$, proving that \hvi assumes a matrix-variate distribution $q(\Wmatr)$, see \cref{fig:covariance} for an illustration of this.

We report the expression for $\Mmatr$, $\Umatr$, and $\Vmatr$ and leave the full derivation to the Supplement.
For the mean, we have $\Mmatr = \Smatr_1\Hmatr\diag(\muvect) \Hmatr\Smatr_2$, whereas
for $\Umatr$ and $\Vmatr$, we have:
\begin{align}
    \Umatr^{1/2} = \Smatr_1 \Hmatr \Tmatr_2  \quad \text{and}   \quad  \Vmatr^{1/2} = \frac{1}{\sqrt{\Tr(\Umatr)}} \Smatr_2 \Hmatr \Tmatr_1,
\end{align}
where each row $i$ of  $\Tmatr_1 \in \R^{D \times D^2}$ is the column-wise vectorization of $(\Sigmamatr^{1/2}_{i,j}(\Hmatr {\Smatr_1})_{i,j'})_{j,j'\le D}$, the matrix $\Tmatr_2$ is defined similarly with $\Smatr_2$ instead of $\Smatr_1$, and $\Tr(\cdot)$ denotes the trace operator.

The mean of the structured matrix-variate posterior assumed by \hvi can span a $D$-dimensional linear subspace within the whole $D^2$-dimensional parameter space, and the orientation is controlled by the matrices $\Smatr_1$ and $\Smatr_2$; more details on this geometric interpretation of \hvi can be found in the Supplement.

Matrix-variate Gaussian posteriors for variational inference have been introduced in \cite{Louizos2016}; however, assuming full covariance matrices $\Umatr$ and $\Vmatr$ is memory and computationally intensive (quadratic and cubic in $D$, respectively).
\hvi captures covariances across weights (see \cref{fig:covariance}), while keeping memory requirements linear in $D$ and complexity log-linear in $D$.

\subsection{Reparameterizations in \hvi for Stochastic Optimization}
\label{sec:local-reparam}

The so-called {\em reparameterization trick} \citep{Kingma14} is a standard way to make the variational lower bound in \cref{eq:lowerbound} a deterministic function of the variational parameters, so as to be able to carry out gradient-based optimization despite the stochasticity of the objective.
Considering input vectors $\hvect_i$ to a given layer, an improvement over this approach is to consider the distribution of the product $\Wmatr \hvect_i$.
This is also known as the {\em local reparameterization trick} \citep{Kingma2015}, and it reduces the variance of stochastic gradients in the optimization, thus improving convergence.
The product $\Wmatr \hvect_i$ follows the distribution $\norm(\mvect,\Amatr\Amatr^\T)$ \citep{Gupta1999}, with
\begin{align}
    \mvect = \Smatr_1 \Hmatr  \diag(\muvect) \Hmatr \Smatr_2 \hvect_i \text{,} \quad \text{and} \quad
    \Amatr = \Smatr_1 \Hmatr  \diag( \Hmatr \Smatr_2 \hvect_i)  \Sigmamatr^{1/2}.
\end{align}
A sample from this distribution can be efficiently computed thanks to the Walsh-Hadamard transform as: $
    \overline\Wmatr(\muvect) \hvect_i + \overline\Wmatr(\Sigmavect^{1/2}\varepsilonvect)\hvect_i \text{,}
$
with $\overline\Wmatr$ a linear matrix-valued function $\overline\Wmatr(\uvect) =  \Smatr_1 \Hmatr  \diag(\uvect) \Hmatr \Smatr_2$.
\subsection{Alternative Structures and Comparison with Tensor Factorization}
\label{sec:comperison-tensorfact}
\begin{figure*}[b]
    \scriptsize
    \setlength\figureheight{.24\textwidth}
    \setlength\figurewidth{.35\textwidth}
    \pgfplotsset{every axis title/.append style={yshift=-1ex}}
    \pgfplotsset{every x tick label/.append style={font=\fontsize{2}{4}\selectfont}}
    \pgfplotsset{every y tick label/.append style={font=\fontsize{2}{4}\selectfont}}
    \begin{minipage}[t]{.6\textwidth}
        \vspace{0ex}\tiny
        \includegraphics{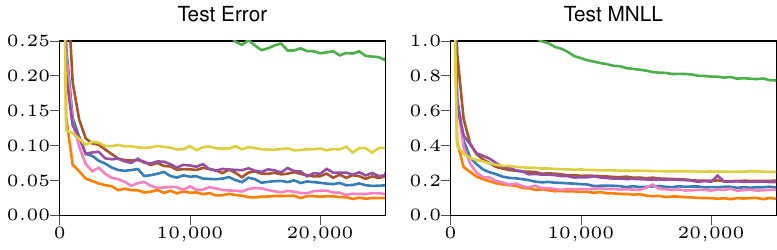}
        \captionof{figure}{Ablation study of different structures for the parameterization of the weights distribution. Metric: test \errate and test \mnll with different structures for the weights. Benchmark on \name{drive} with a $2\times64$ network.}
        \label{fig:different_structures}
    \end{minipage}
    \hfill
    \begin{minipage}[t]{.37\textwidth}
        \tiny
        \vspace{0ex}
        \captionof{table}{List of alternative structures and test performance on \name{drive} dataset.}
        \begin{center}
            {\sc\centering
                \vspace{-5ex}
                \definecolor{color5}{rgb}{0.596078431372549,0.305882352941176,0.63921568627451}
\definecolor{color1}{rgb}{1,0.498039215686275,0}
\definecolor{color3}{rgb}{0.968627450980392,0.505882352941176,0.749019607843137}
\definecolor{color0}{rgb}{0.215686274509804,0.494117647058824,0.72156862745098}
\definecolor{color2}{rgb}{0.301960784313725,0.686274509803922,0.290196078431373}
\definecolor{color6}{rgb}{0.870588235294118,0.811764705882353,0.247058823529412}
\definecolor{color4}{rgb}{0.650980392156863,0.337254901960784,0.156862745098039}
\begin{tabular}{lrr}
    \toprule
                                                                                                       & \multicolumn{2}{c}{test}                \\
    {}                                                                                                 & error                    & mnll         \\
    model                                                                                              &                          &              \\
    \midrule
    \rowcolor{color6!30}mcd                                                                            & $0.097$                  & $0.249$      \\
    \rowcolor{color2!30!white}$\Gmatr\Hmatr$                                                           & $0.226$                  & $0.773$      \\
    \rowcolor{color0!30!white}$\Smatr_{\mathrm{var}}\Hmatr\Gmatr\Hmatr$                                & $0.043$                  & $0.159$      \\
    \rowcolor{color5!30!white}$\Smatr_{1,\mathrm{var}}\Hmatr\Gmatr\Hmatr\Smatr_{2,\mathrm{var}}\Hmatr$ & $0.061$                  & $0.190$      \\
    \rowcolor{color4!30!white}$\Smatr_{\mathrm{opt}}\Hmatr\Gmatr\Hmatr$                                & $0.054$                  & $0.199$      \\
    \rowcolor{color3!30!white}$\Smatr_{1,\mathrm{opt}}\Hmatr\Gmatr\Hmatr\Smatr_{2,\mathrm{opt}}\Hmatr$ & $0.031$                  & $0.146$      \\
    \rowcolor{color1!30!white}$\Smatr_{1,\mathrm{opt}}\Hmatr\Gmatr\Hmatr\Smatr_{2,\mathrm{opt}}$ (\hvi)                                & $\bm{0.026}$             & $\bm{0.094}$ \\
    \bottomrule
\end{tabular}

            }
        \end{center}
        \label{tab:different_structures}
        \vspace{1ex}\par
        \tiny{Colors are coded to match the ones used in the adjacent Figure}
    \end{minipage}
\end{figure*}

The choice of the parameterization of $\Wmatr$ in \hvi leaves space to several possible alternatives, which we compare in \cref{tab:different_structures}.
For all of them, $\Gmatr$ is learned variationally and the remaining diagonal $\Smatr_i$ (if any) are either optimized or treated variationally (Gaussian mean-field).
\cref{fig:different_structures} shows the behavior of these alternatives when applied to a $2\times64$ network with \relu activations.
With the exception of the simple and highly constrained alternative $\Gmatr\Hmatr$, all parameterizations are converging quite easily and the comparison with \mcd shows that indeed the proposed \hvi performs better both in terms of \errate and \mnll.
\begin{figure}[t]
    \vspace{0ex}
    \tiny
    \setlength\figureheight{.23\textwidth}
    \setlength\figurewidth{.32\textwidth}
    \begin{minipage}{.55\textwidth}
        \centering
        \pgfplotsset{every axis title/.append style={yshift=-1ex}}
        \pgfplotsset{every x tick label/.append style={font=\fontsize{2}{4}\selectfont}}
        \pgfplotsset{every y tick label/.append style={font=\fontsize{2}{4}\selectfont}}
        \includegraphics{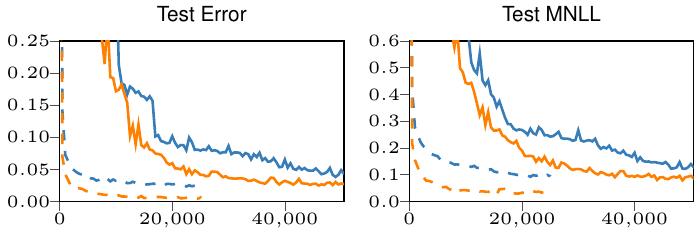}
        \includegraphics{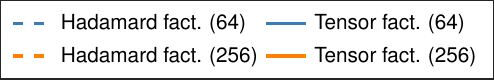}
        \captionof{figure}{Comparison between Hadamard factorization in \hvi and tensor factorization. The number in the parenthesis is the hidden dimension. Plot is w.r.t. iterations rather then time to avoid implementation artifacts. The dataset used is \name{drive}.}
        \label{fig:tensor_fact}
    \end{minipage}
    \hfill
    \begin{minipage}{.4\textwidth}
        \begin{algorithm}[H]
            \scriptsize
            \SetKwFunction{FMain}{SetupDimensions}
            \SetKwProg{Fn}{Function}{:}{}
            \Fn{\FMain{$\Din,\Dout$}}{
                $\text{next power} \gets 2 ^ {\ceil{\log_2\Din}}$\;
                \eIf{next power == $2\Din$}{
                    $\text{padding} \gets 0$\;
                }{
                    $\text{padding} = \text{next power} - \Din$\;
                    $\Din\gets \text{next power}$\;
                }
                stack, remainder = divmod($\Dout$, $\Din$)\;
                \If{remainder != 0} {
                    $\text{stack} \gets \text{stack} + 1$\;
                    $\Dout \gets \Din \times \text{stack}$\;
                }
                \Return{$\Din$, $\Dout$, padding, stack}
            }
            \caption{Setup dimensions for non-squared matrix}
            \label{alg:setup_dimensions}
        \end{algorithm}

    \end{minipage}
\end{figure}
\hvi is effectively imposing a factorization of $\Wmatr$, where parameters are either optimized or treated variationally.
Tensor decompositions for \dnns and \cnns have been proposed in \citep{Novikov2015}; here $\Wmatr$ is decomposed into $k$ small matrices (tensor cores), such that
$ \Wmatr = {\Wmatr}_1 {\Wmatr}_2\cdots {\Wmatr}_{k}\,,
$ where each $\Wmatr_i$ has dimensions $r_{i-1}\times r_i$ (with $r_1 = r_k = D$).
We adapt this idea to make a comparison with \hvi. In order to match the space and time complexity of \hvi, assuming $\{r_i = R|\forall i=2,\dots,k-1\}$, we set:
$ R \propto \log_2 D   \quad \text{and} \quad   K \propto \frac{D}{(\log_2 D)^2}\,.
$ Also, to match the number of variational parameters, all internal cores ($i=2,\dots,k-1$) are learned with fully factorized Gaussian posterior, while the remaining are optimized (see \cref{tab:complexity}).
Given the same asymptotic complexity, \cref{fig:tensor_fact} reports the results of this comparison again on a 2 hidden layer network.
Not only \hvi can reach better solutions in terms of test performance, but optimization is also faster.
We speculate that this is attributed to the redundant variational parameterization induced by the tensor cores, which makes the optimization landscapes highly multi-modal, leading to slow convergence.
\subsection{Extensions}

\paragraph{Concatenating or Reshaping Parameters for \hvi}
For the sake of presentation, so far we have assumed $\Wmatr \in \R^{D \times D}$ with $D = 2^d$, but we can easily extend \hvi to handle parameters of any shape $\Wmatr \in \R^{\Dout \times \Din}$.
One possibility is to use \hvi with a large $D \times D$ matrix with $D = 2^d$, such that a subset of its elements represent $\Wmatr$.
Alternatively, a suitable value of $d$ can be chosen so that $\Wmatr$ is a concatenation by row/column of square matrices of size $D = 2^d$, padding if necessary (\cref{alg:setup_dimensions} shows this case).

When one of the dimensions is equal to one so that the parameter matrix is a vector ($\Wmatr = \wvect \in \R^D$), this latter approach is not ideal, as \hvi would fall back on mean-field \vi.
\hvi can be extended to handle these cases efficiently by reshaping the parameter vector into a matrix of size $2^d$ with suitable $d$, again by padding if necessary.
Thanks to the reshaping, \hvi uses $\sqrt{D}$ parameters to model a posterior over $D$, and allows for computations in $\bigO(\sqrt{D}\log D)$ rather than $D$.
This is possible by reshaping the vector that multiplies the weights in a similar way.
In the Supplement, we explore this idea to infer parameters of Gaussian processes linearized using large numbers of random features.

\paragraph{Normalizing Flows}
Normalizing flows \citep[\nf,][]{Rezende2015} are a family of parameterized distributions that allow for flexible approximations.
In the general setting, consider a set of invertible, continuous and differentiable functions $f_k:\R^D\rightarrow\R^D$ with parameters $\lambdavect_k$.
Given $\zvect_0 \sim q_0(\zvect_0)$, $\zvect_0$ is transformed with a chain of $K$ flows to $\zvect_K=(f_K\circ\cdots\circ f_1)(\zvect_0)$.
The variational lower bound slightly differs from \cref{eq:lowerbound} to take into account the determinant of the Jacobian of the transformation, yielding a new variational objective as follows:
\begin{align}
    \Exp_{q_0} \left[ \log p(\Ymatr|\Xmatr, \Wmatr) \right] - \KL\{q_0(\zvect_0)||p(\zvect_K)\}  
    + \Exp_{q_0(\zvect_0)}\left[\sum\nolimits_{k=1}^K \log \left|\det \frac{\partial f_k(\zvect_{k-1};\lambdavect_k)}{\partial \zvect_{k-1}} \right|\right]\,.
\end{align}
Setting the initial distribution $q_0$ to a fully factorized Gaussian $\N(\zvect_0|\muvect, \sigmavect\Ivect)$ and assuming a Gaussian prior on the generated $\zvect_K$, the \KL term is analytically tractable.
The tranformation $f$ is generally chosen to allow for fast computation of the determinant of the Jacobian.
The parameters of the initial density $q_0$ as well as the flow parameters $\lambdavect$ are optimized.
In our case, we consider $q_K$ as a distribution over the elements of $\gvect$.
This approach increases the flexibility of the form of the variational posterior in \hvi, which is no longer Gaussian, while still capturing covariances across weights.
This is obtained at the expense of losing the possibility of employing the local reparameterization trick.
In the following Section, we will use {\em planar flows} \citep{Rezende2015}.
Although this is a simple flow parameterization, a planar flow requires only $\bigO(D)$ parameters and thus it does not increase the time/space complexity of \hvi.
More complex alternatives can be found in \citep{Berg2018, Kingma2016, Louizos2017}.

\section{Experiments}
\label{sec:experiments}
In this Section we will provide experimental evaluations of our proposal, with experiments ranging from regression on classic benchmark datasets to image classification with large-scale convolutional neural networks. We will also comment on the computational efficiency and some potential limitation of our proposal.

\subsection{Toy example}
\begin{figure}[!t]
    \tiny
    \centering
    \setlength\figureheight{.22\textwidth}
    \setlength\figurewidth{.27\textwidth}
    \pgfplotsset{every x tick label/.append style={font=\fontsize{2}{4}\selectfont}}
    \pgfplotsset{every y tick label/.append style={font=\fontsize{2}{4}\selectfont}}
    \pgfplotsset{every axis title/.append style={yshift=-2ex}}
    \begin{minipage}[t]{.78\textwidth}
        \includegraphics{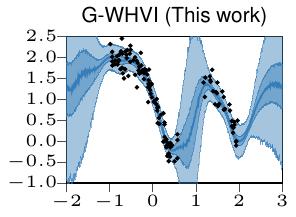}\hspace{-2ex}\includegraphics{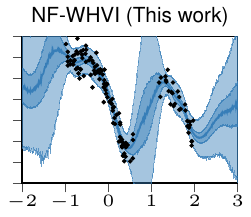}\hspace{-2ex}\includegraphics{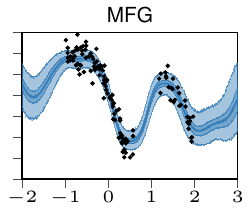}\hspace{-2ex}\includegraphics{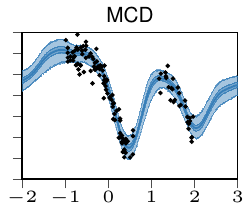}\end{minipage}
    \hfill\unskip\ \vrule\ \begin{minipage}[t]{.2\textwidth}
        \includegraphics{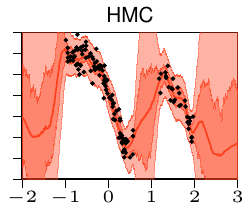}
        \end{minipage}
    \caption{Regression example trained using \hvi with Gaussian vector (1541 param.) and with planar normalizing flow (10 flows for a total of 4141 param.), \name{mfg} (35k param.) and Monte Carlo dropout (\mcd) (17k param.). The two shaded areas represent the 95th and the 75th percentile of the predictions. As ``ground truth'', we also show the predictive posterior obtained by running \name{sghmc} on the same model ($R < 1.05$, \citep{Gelman2004}).}
    \label{fig:1d_regression}
\end{figure}
We begin our experimental validation with a 1D-regression problem. 
We generated a 1D toy regression problem with 128 inputs sampled from $\mathcal{U}[-1, 2]$, and removed $20\%$ inputs on a predefined interval;
targets are noisy realizations of a random function (noise variance $\sigma^2 = \exp(-3)$). We model these data using a \dnn with 2 hidden layers of 128 features and cosine activations.
We test four models: mean-field Gaussian \vi (\name{mfg}), Monte Carlo dropout \citep[\mcd,][]{Gal16} with dropout rate $0.4$ and two variants of \hvi\xspace -- \name{g-whvi} with Gaussian posterior and \name{nf-whvi} with planar flows (10 planar flows).
We also show the free form posterior obtained by running a \name{mcmc} algorithm, \name{sghmc} in this case \citep{Cheni2014, Springenberg2016}, for several thousands steps.
As \cref{fig:1d_regression} shows, \hvi offers a sensible modeling of the uncertainty on the input domain, whereas \name{mfg} and \mcd seem to be slightly over-confident.

\begin{table*}[b]
    \sc
    \scriptsize
    \centering
    \caption{Test \rmse and test \mnll for regression datasets. Results in the format ``\textit{mean~(std)}''}
    \vspace{-2ex}
    \label{tab:regression_bnn}
    \tiny
\renewcommand{\tabcolsep}{1.5ex}
\begin{tabular}{l||rrrr|rrrr}
    \toprule
    {}         & \multicolumn{4}{r|}{{test error}}   & \multicolumn{4}{r}{{test mnll}}                                                                                                                                                                                                                                    \\
    model      & mcd                                 & mfg                             & nng                                 & whvi                                 & mcd                                 & mfg                             & nng                                  & whvi                                 \\
    dataset    &                                     &                                 &                                     &                                      &                                     &                                 &                                      &                                      \\
    \midrule
    boston     & $3.91$ \scalebox{.8}{$(0.86)$}      & $4.47$ \scalebox{.8}{$(0.85)$}  & $3.56$ \scalebox{.8}{$(0.43)$}      & $\bm{3.14}$  \scalebox{.8}{$(0.71)$} & $ 6.90$ \scalebox{.8}{$(2.93)$}     & $ 2.99$ \scalebox{.8}{$(0.41)$} & $\bm{2.72}$ \scalebox{.8}{$(0.09)$}  & $4.33$  \scalebox{.8}{$(1.80)$}      \\
    concrete   & $5.12$ \scalebox{.8}{$(0.79)$}      & $8.01$ \scalebox{.8}{$(0.41)$}  & $8.21$ \scalebox{.8}{$(0.55)$}      & $\bm{4.70}$  \scalebox{.8}{$(0.72)$} & $ 3.20$ \scalebox{.8}{$(0.36)$}     & $ 3.41$ \scalebox{.8}{$(0.05)$} & $3.56$  \scalebox{.8}{$(0.08)$}      & $ \bm{3.17}$ \scalebox{.8}{$(0.37)$} \\
    energy     & $2.07$ \scalebox{.8}{$(0.11)$}      & $3.10$ \scalebox{.8}{$(0.14)$}  & $1.96$ \scalebox{.8}{$(0.28)$}      & $\bm{0.58}$  \scalebox{.8}{$(0.07)$} & $ 4.15$ \scalebox{.8}{$(0.15)$}     & $ 4.91$ \scalebox{.8}{$(0.09)$} & $2.11$  \scalebox{.8}{$(0.12)$}      & $ \bm{2.00}$ \scalebox{.8}{$(0.60)$} \\
    kin8nm     & $0.09$ \scalebox{.8}{$(0.00)$}      & $0.12$ \scalebox{.8}{$(0.00)$}  & $\bm{0.07}$ \scalebox{.8}{$(0.00)$} & $0.08$  \scalebox{.8}{$(0.00)$}      & $-0.87$ \scalebox{.8}{$(0.02)$}     & $-0.83$ \scalebox{.8}{$(0.02)$} & $\bm{-1.19}$ \scalebox{.8}{$(0.04)$} & $\bm{-1.19}$ \scalebox{.8}{$(0.04)$} \\
    naval      & $0.30$ \scalebox{.8}{$(0.30)$}      & $0.01$ \scalebox{.8}{$(0.00)$}  & $\bm{0.00}$ \scalebox{.8}{$(0.00)$} & $0.01$  \scalebox{.8}{$(0.00)$}      & $-1.00$ \scalebox{.8}{$(2.27)$}     & $-6.23$ \scalebox{.8}{$(0.01)$} & $\bm{-6.52}$ \scalebox{.8}{$(0.09)$} & $-6.25$ \scalebox{.8}{$(0.01)$}      \\
    powerplant & $\bm{3.97}$ \scalebox{.8}{$(0.14)$} & $4.52$ \scalebox{.8}{$(0.13)$}  & $4.23$ \scalebox{.8}{$(0.09)$}      & $4.00$ \scalebox{.8}{$(0.12)$}       & $ 2.74$ \scalebox{.8}{$(0.05)$}     & $ 2.83$ \scalebox{.8}{$(0.03)$} & $2.86$  \scalebox{.8}{$(0.02)$}      & $\bm{2.71}$ \scalebox{.8}{$(0.03)$}  \\
    protein    & $\bm{4.23}$ \scalebox{.8}{$(0.10)$} & $4.93$ \scalebox{.8}{$(0.11)$}  & $4.57$ \scalebox{.8}{$(0.47)$}      & $4.36$ \scalebox{.8}{$(0.11)$}       & $\bm{2.76}$ \scalebox{.8}{$(0.02)$} & $ 2.92$ \scalebox{.8}{$(0.01)$} & $2.95$  \scalebox{.8}{$(0.12)$}      & $2.79$ \scalebox{.8}{$(0.01)$}                      \\
    yacht      & $1.90$ \scalebox{.8}{$(0.54)$}      & $7.01$ \scalebox{.8}{$(1.22)$}  & $5.16$ \scalebox{.8}{$(1.48)$}      & $\bm{0.69}$ \scalebox{.8}{$(0.16)$}  & $2.95$ \scalebox{.8}{$(1.27)$}      & $ 3.38$ \scalebox{.8}{$(0.29)$} & $3.06$  \scalebox{.8}{$(0.27)$}      & $\bm{1.80}$ \scalebox{.8}{$(1.01)$}                 \\
    \bottomrule
\end{tabular}

\end{table*}

\subsection{Bayesian Neural Networks}
We conduct a series of comparisons with state-of-the-art \vi schemes for Bayesian \dnns; see the Supplement for the list of data sets used in the experiments.
We compare \hvi with \mcd and \name{nng} \citep[\name{noisy-kfac},][]{Zhang2018}.
\mcd draws on a formal connection between dropout and \vi with Bernoulli-like posteriors, while the more recent \name{noisy-kfac} yields a matrix-variate Gaussian distribution using noisy natural gradients.
To these baselines, we also add the comparison with mean field Gaussian (\name{mfg}).
In \hvi, the last layer assumes a fully factorized Gaussian posterior.

Data is randomly divided into 90\%/10\% splits for training and testing eight times.
We standardize the input features $\xvect$ while keeping the targets $\yvect$ unnormalized.
Differently from the experimental setup in \citep{Louizos2016, Zhang2018, Lobato2015}, we use the same architecture regardless of the size of the dataset.
Futhermore, to test the efficiency of \hvi in case of over-parameterized models, we set the network to have two hidden layers and 128 features with \relu activations (as a reference, these models are $\sim$20 times bigger than the usual setup, which uses a single hidden layer with 50/100 units).

We report the test \rmse and the average predictive test negative log-likelihood (\mnll) in \cref{tab:regression_bnn}.
On the majority of the datasets, \hvi outperforms \mcd and \name{noisy-kfac}.

\setlength\figureheight{.24\textwidth}
\setlength\figurewidth{.43\textwidth}
\begin{minipage}{.49\textwidth}
    Futhermore, we study how the test \mnll varies with the number of hidden units in a 2-layered network.
    As \cref{fig:features_vs_mnll} shows, \hvi behaves well while competitive methods struggle.
    Empirically, these results demonstrate the value of \hvi, which offers a competitive parameterization of a matrix-variate Gaussian posterior while requiring log-linear time in $D$.
We refer the Reader to the Supplement for additional details on the experimental setup and for the benchmark with the classic architectures.
\end{minipage}
\hfill
\begin{minipage}{.49\textwidth}
    \tiny
    \centering
    \includegraphics{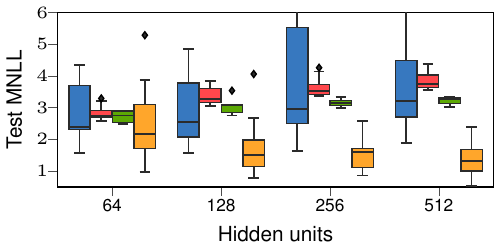}
    \definecolor{color0}{rgb}{0.215686274509804,0.470588235294118,0.749019607843137}  \definecolor{color3}{rgb}{1,0.650980,0.1686274}                           \definecolor{color2}{rgb}{0.360784313725490,0.662745098039216,0.0156862745098039}   \definecolor{color1}{rgb}{1,0.27843137254902,0.298039215686275}    \tikzexternaldisable
    \captionof{figure}{{
    Comparison of the test \mnll as a function of the number of hidden units for
    \mcd
    (\protect\tikz \protect\draw[thick, color=color0, fill=color0] plot[mark=*, line width=5pt, mark options={scale=1.3}] (0,0.5);),
    \name{mfg}
    (\protect\tikz \protect\draw[thick, color=color1, fill=color1] plot[mark=*, line width=5pt, mark options={scale=1.3}] (0,0.5);),
    \name{nng}
    (\protect\tikz \protect\draw[thick, color=color2, fill=color2] plot[mark=*, line width=5pt, mark options={scale=1.3}] (0,0.5);) and
    \hvi
    (\protect\tikz \protect\draw[thick, color=color3, fill=color3] plot[mark=*, line width=5pt, mark options={scale=1.3}] (0,0.5);).
    The dataset used is \name{yacht}.
    }}
    \tikzexternalenable
    \label{fig:features_vs_mnll}
\end{minipage}

\subsection{Bayesian Convolutional Neural Networks}
\begin{figure}[b]
    \begin{minipage}[t]{.49\textwidth}
        \sc
        \scriptsize
        \centering
        \vspace{0pt}
        \captionof{table}{Test performance of different Bayesian \cnns.}
        \label{tab:cnn_results}
        \renewcommand{\arraystretch}{.5}
\begin{tabular}{rl|rrr}
    \toprule
             & {\cifart}  & test error & test mnll \\
    \midrule
    vgg16    & mfg        & $16.82\%$  & $0.6443$  \\
             & mcd        & $21.47\%$  & $0.8213$  \\
             & nng        & $15.21\%$  & $\mathbf{0.6374}$  \\
             & \hvi       & $\mathbf{12.85\%}$  & $0.6995$  \\
    \midrule
    alexnet  & \mcd       & $13.30\%$  & $0.9590$  \\
             & \name{nng} & $20.36\%$  & --        \\ & \hvi       & $13.56\%$  & $\mathbf{0.6164}$  \\  & \nf-\hvi   & $\mathbf{12.72\%}$  & $0.6596$  \\  \midrule
    resnet18 & \mcd       & $\mathbf{10.71\%}$  & $0.8468$  \\
             & \name{nng} & --  & --        \\ & \hvi       & $11.46\%$  & $0.5513$  \\
             & \nf-\hvi   & $11.42\%$  & $\mathbf{0.4908}$  \\ \bottomrule
\end{tabular}

    \end{minipage}
    \hfill
    \begin{minipage}[t]{.49\textwidth}
        \vspace{-7.5pt}
        \tiny
        \centering
        \pgfplotsset{every axis title/.append style={yshift=-2ex}}
        \setlength\figureheight{.45\textwidth}
        \pgfplotsset{every x tick label/.append style={font=\fontsize{2}{4}\selectfont}}
        \pgfplotsset{every y tick label/.append style={font=\fontsize{2}{4}\selectfont}}
        \setlength\figurewidth{.9\textwidth}
        \includegraphics{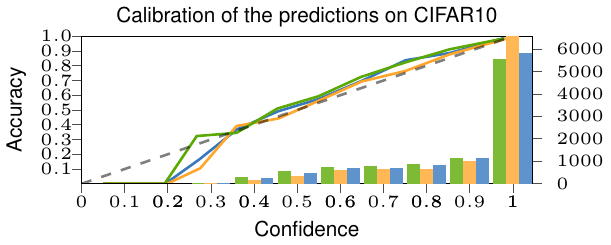}\\
        \includegraphics{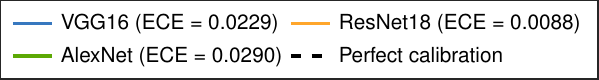}
        \captionof{figure}{Reliability diagram and expected calibration error (\ece) of \vgg, \alexnet and \resnet with \hvi \citep{DeGroot1983,Niculescu-Mizil05,Naeini15}.}
        \label{fig:calibration_cnn}
    \end{minipage}
\end{figure}

We continue the experimental evaluation of \hvi by analyzing its performance on \cnns.
For this experiment, we replace all fully-connected layers in the \cnn with the \hvi parameterization, while the convolutional filters are treated variationally using \mcd.
In this setup, we fit  \vgg~\citep{Simonyan14}, \alexnet~\citep{Krizhevsky2012} and \resnet-18~\citep{He16} on \cifart.
Using \hvi, we can reduce the number of parameters in the linear layers without affecting neither test performance nor calibration properties of the resulting model, as shown in \cref{fig:calibration_cnn} and \cref{tab:cnn_results}.
For \alexnet and \resnet we also try our variant of \hvi with \nf.
Even though we lose the benefits of the local reparameterization, the higher flexibility of normalizing flows allows the model to obtain better test performance with respect to the Gaussian posterior.
This can be improved even further using more complex families of normalizing flows \citep{Rezende2015, Berg2018, Kingma2016, Louizos2017}.
With \hvi, \alexnet and its original $\sim$23.3\name{m} parameters is reduced to just $\sim$2.3\name{m} (9.9\%) when using \name{g}-\hvi and to $\sim$2.4\name{m} (10.2\%) with \hvi and 3 planar flows.

\paragraph{\textsc{whvi} for convolutional filters}
By observing that the convolution can be written as matrix multiplication (once filters are reshaped in 2D), we also extended \hvi for convolutional layers.

\begin{minipage}{.38\textwidth}
    We observe though that in this case resulting models had too few parameters to obtain any interesting results.
    For \alexnet, we obtained a model with just 189\name{k} parameters, which corresponds to a sparsity of 99.2\% with respect of the original model.
    As a reference, \citet{Wen2016} was able to reach sparsity only up to 60\% in the convolutional layers without impacting performance.
\end{minipage}\hfill\begin{minipage}{.6\textwidth}
    \tiny
    \centering
    \pgfplotsset{every axis title/.append style={yshift=-2ex}}
    \setlength\figureheight{3.25cm}
    \setlength\figurewidth{5cm}
    \pgfplotsset{every x tick label/.append style={font=\fontsize{2}{4}\selectfont}}
    \pgfplotsset{every y tick label/.append style={font=\fontsize{2}{4}\selectfont}}
    \includegraphics{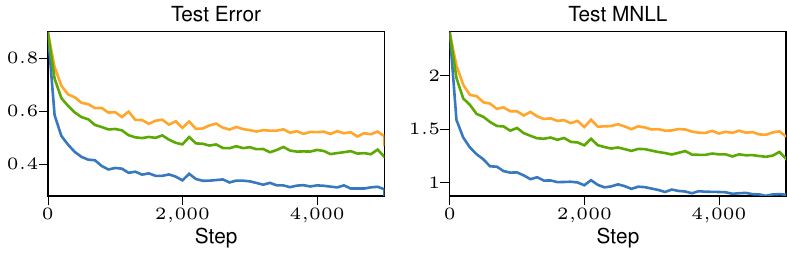}
    \\
    \sffamily
    \tikzexternaldisable
    \setlength{\tabcolsep}{3pt}
    \definecolor{color2}{rgb}{0.3019,0.6862,0.2901}
    \definecolor{color1}{rgb}{1,0.4980,0}
    \definecolor{color0}{rgb}{0.2156,0.4941,0.7215}
    \begin{tabular}{|cll|}
        \hline &                                                                             &                             \\[-5pt]
        {\protect\tikz[baseline=-.75ex]\protect\draw[thick, color=color0, fill=color0, line width=.75pt] plot[] (-.0, 0)--(.35,0);}
               & {W\textsubscript{conv} with MCD -- W\textsubscript{lin} with WHVI}          & Error = 0.281, MNLL = 0.882 \\
        {\protect\tikz[baseline=-.75ex]\protect\draw[thick, color=color2, fill=color2, line width=.75pt] plot[] (-.0, 0)--(.35,0);}
               & {W\textsubscript{conv} with WHVI -- W\textsubscript{lin} with WHVI}         & Error = 0.427, MNLL = 1.223 \\
        {\protect\tikz[baseline=-.75ex]\protect\draw[thick, color=color1, fill=color1, line width=.75pt] plot[] (-.0, 0)--(.35,0);}
               & {W\textsubscript{conv} low-rank with MCD -- W\textsubscript{lin} with WHVI} & Error = 0.469, MNLL = 1.434 \\[1pt]
        \hline
    \end{tabular}
    \tikzexternalenable
    \captionof{figure}{Inference of convolutional filters (dataset: \name{cifar10}).}
    \label{fig:param-cnn}
\end{minipage}

To study this behavior in details, we take a simple \name{cnn} with two convolutional layers and one linear layer (\cref{fig:param-cnn}).
We see that the combination of \mcd and \hvi performs very well in terms of convergence and test performance, while the use of \hvi on the convolutional filters brings an overall degradation of the performance.
Interestingly, though, we also observe that \mcd with the same number of parameters as for \hvi (referred to as low-rank \mcd) performs even worse than the baseline: this once again confirms the parameterization of \hvi as an efficient alternative.

\subsection{Comments on computational efficiency}

\hvi builds his computational efficiency on the Fast Walsh-Hadamard Transform (\name{fwht}), which allows one to cut the complexity of a $D$-dimensional matrix-vector multiplication from a naive $\bigO(D^2)$ to $\bigO(D\log D)$.
To empirically validate this claim, we extended \pytorch \citep{Paszke2017} with a custom \name{c++}/\name{cuda} kernel which implements a batched-version of the \name{fwht}.
The workstation used is equipped with two Intel Xeon \textsc{cpu}s, four NVIDIA Tesla P100 and 512~GB of RAM. Each experiment is carried out on a \name{gpu} fully dedicated to it.
The \name{nng} algorithm is implemented in \tensorflow \footnotemark \xspace while the others are written in \pytorch.
\footnotetext[1]{\scriptsize\href{https://github.com/gd-zhang/noisy-K-FAC}{\texttt{github.com/gd-zhang/noisy-K-FAC}} --- \href{https://github.com/pomonam/NoisyNaturalGradient}{\texttt{github.com/pomonam/NoisyNaturalGradient}}}

\setlength\figureheight{.22\textwidth}
\setlength\figurewidth{.28\textwidth}
\begin{minipage}{.49\textwidth}
    We made sure to fully exploit all parallelization opportunities in the competiting methods and ours; we believe that the timings are not severely affected by external factors other than the actual implementation of the algorithms.
    The box-plots in \cref{fig:inference_time} report the time required to sample and infer the carry out inference on the test set on two regression datasets as a function of the number of hidden units in a two-layer \dnn .
    We speculate that the poor performance of \name{nng} is due to the inversion of the approximation to the Fisher matrix, which scales cubically in the number of units.
    \end{minipage}
\hfill
\begin{minipage}{.49\textwidth}
    \tiny
    \centering
    \pgfplotsset{every y tick label/.append style={font=\fontsize{1}{3}\selectfont}}
    \pgfplotsset{every axis title/.append style={yshift=-2ex}}
    \includegraphics{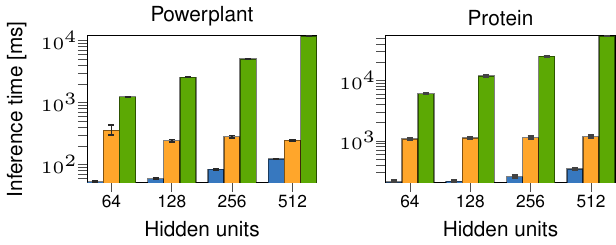}
    \includegraphics{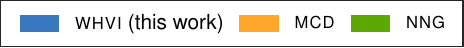}
    \captionof{figure}{Inference time on the test set with 128 batch size and 64 Monte Carlo samples. Experiment repeated 100 times. Additional datasets available in the Supplement.
        }
    \label{fig:inference_time}
\end{minipage}

\setlength\figureheight{.3\textwidth}
\setlength\figurewidth{.45\textwidth}
\begin{minipage}{.49\textwidth}
    Similar behavior can also be observed for Bayesian \cnns.
    In \cref{fig:gpustats}, we analyze the energy consumption required to sample from the converged model and predict on the test set of \cifart with \alexnet using \hvi and \mcd.
    The regularity of the algorithm for computing the \name{fwht} and its reduced memory footprint result on an overall higher utilization of the \name{gpu},
    $85\%$ for \hvi versus $\sim70\%$ for \mcd.
    This translates into an increase of energy efficiency up to $33\%$ w.r.t \mcd, despite being $51\%$ faster.
    \paragraph{Additional results and insights}
    We refer the reader to the Supplement for an extended version of the results, including new applications of \hvi to \gps.
\end{minipage}
\hfill
\begin{minipage}{.49\textwidth}
    \centering\tiny
    \pgfplotsset{every axis title/.append style={yshift=-2ex}}
    \includegraphics{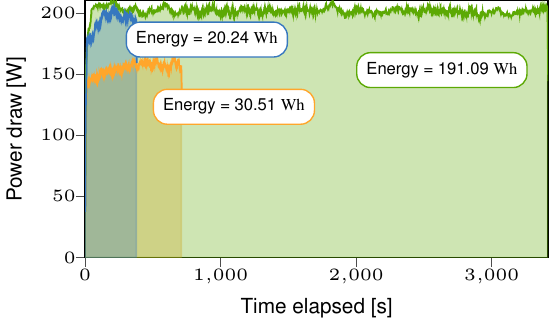}\\
    
    \definecolor{color1}{rgb}{1,0.650980,0.1686274}
    \definecolor{color2}{rgb}{0.360784313725490,0.662745098039216,0.0156862745098039}
    \definecolor{color0}{rgb}{0.215686274509804,0.470588235294118,0.749019607843137}
    \tikzexternaldisable
    \captionof{figure}{{
                Power profiling during inference on the test set of \cifart with \alexnet and \hvi
                (\protect\tikz \protect\draw[thick, color=color0, fill=color0] plot[mark=*, line width=5pt, mark options={scale=1.3}] (0,0.5);), \mcd
                (\protect\tikz \protect\draw[thick, color=color1, fill=color1] plot[mark=*, line width=5pt, mark options={scale=1.3}] (0,0.5);)
                and \name{nng}~
                (\protect\tikz \protect\draw[thick, color=color2, fill=color2] plot[mark=*, line width=5pt, mark options={scale=1.3}] (0,0.5);).
                The task is repeated 16 consecutive times and profiling is carried out using the \texttt{nvidia-smi} tool.
            }}
    \tikzexternalenable
    \label{fig:gpustats}
\end{minipage}

\section*{Related Work}
\vspace{-0.15cm}
In the early sections of the paper, we have already briefly reviewed some of the literature on \vi and Bayesian \dnns and \cnns; here we complement the literature by including other relevant works that have connections with \hvi.

Our work takes inspiration from the works on random features for kernel approximation \citep{Rahimi08} and \fastfood \citep{Le13}.
Random feature expansions have had a wide impact on the literature on kernel methods. 
Such approximations have been successfully used to scale a variety of models, such as Support Vector Machines \citep{Rahimi08}, Gaussian processes \citep{LazaroGredilla10} and Deep Gaussian processes \citep{Cutajar17, Gal16}. This has contributed to bridging the gap between Deep \gps and Bayesian \dnns and \cnns \citep{Neal1997, Duvenaud14, Cutajar17, Gal16b}, which is an active area of research which aims to gain a better understanding of deep learning models through the use of kernel methods \citep{Matthews2018, Matthew2018, Garriga2018}.
Structured random features \citep{Le13, Yu16, Bojarski2017} have been also applied to the problem of handling large dimensional convolutional features \citep{Yang2015} and Convolutional \gps \citep{Tran18}.

Bayesian inference on \dnns and \cnns has been research topic of several seminar works \citep[see e.g.~][]{Graves11, Lobato2015, Blundell2015, Gal16, Gal16b}.
Recent advances in \dnns have investigated the effect of over-parameterization and how model compression can be used during or after training \citep{Hubara2016, Louizos2017a, Zhu2018}.
Our current understanding shows that model performance is affected by the network size with bigger and wider neural networks being more resilient to overfit
\citep{Neyshabur2014, Neyshabur2018}.
For variational inference, and Bayesian inference in general, over-parameterization is reflected on over-regularization of the objective, leading the optimization to converge to trivial solutions (posterior equal to prior).
Several works have encountered and proposed solutions to such issue \citep{Higgins2017, Burgess2018,Bowman2016, Sonderby2016, Rossi2018}.
The problem of how to run accurate Bayesian inference on over-parametrized models like \bnn is still an ongoing open question \citep{Wilson2020, Wenzel2020}

\section{Conclusions}
\label{sec:conclusions}

Inspired by the literature on scalable kernel methods, this paper proposed Walsh-Hadamard Variational Inference (\hvi). 
\hvi offers a novel parameterization of the variational posterior, which is particularly attractive for over-parameterized models, such as modern \dnns and \cnns.
\hvi assumes a matrix-variate posterior distribution, which therefore captures covariances across weights.
Crucially, unlike previous work on matrix-variate posteriors for \vi, this is achieved with a light parameterization and fast computations, bypassing the over-regularization issues of \vi for over-parameterized models. 
The large experimental campaign, demonstrates that \hvi is a strong competitor of other variational approaches for such models, while offering considerable speedups.

We are currently investigating other extensions where we capture the covariance between weights across layers, by either sharing the matrix $\Gmatr$ across, or by concatenating all weights into a single matrix which is then treated using \hvi, with the necessary adaptations to handle the sequential nature of computations. 
Finally, we are looking into deriving error bounds when using \hvi to approximate a generic matrix distribution; as preliminary work, in a numerical study in the supplement we show that the weights induced by \hvi can approximate reasonably well any arbitrary weight matrix, showing a consistent behavior w.r.t. increasing dimensions $D$.

\FloatBarrier
\section*{Broader Impact}
Bayesian inference for Deep Neural Networks (\dnns) and Convolutional Neural Networks (\cnns) offers attractive solutions to many problems where one needs to combine the flexibility of these deep models with the possibility to accurately quantify uncertainty in predictions and model parameters. 
This is of fundamental importance in an increasingly large number of applications of machine learning in society where uncertainty matters, and where calibration of the predictions and resilience to adversarial attacks are desirable.

Due to the intractability of Bayesian inference for such models, one needs to resort to approximations. 
Variational inference (\vi) gained popularity before long the deep learning revolution, which has seen a considerable interest in the application of \vi to \dnns and \cnns in the last decade. 
However, \vi is still under appreciated in the deep learning community because it comes with a higher computational cost for optimization, sampling, storage and inference. 
With this work, we offer a novel solution to this problem to make \vi truly scalable in each of its parts (parameterization, sampling and inference).

Our approach is inspired by the literature on kernel methods, and we believe that this cross-fertilization will enable further contributions in both communities. 
In the long term, our work will make it possible to accelerate training/inference of Bayesian deep models, while reducing their storage requirements. 
This will complement Bayesian compression techniques to facilitate the deployment of Bayesian deep models onto \name{fpga}, \name{asic} and embedded processors.

\begin{ack}
The Authors would like to thanks Dino Sejdinovic for the insightful discussion on tensor decomposition, which resulted in the comparison in \cref{sec:comperison-tensorfact}.
SR would like to thank Pietro Michiardi for allocating significant resources to our experimental campaign on the Zoe cloud computing platform \citep{pace2017}.
MF gratefully acknowledges support from the AXA Research Fund and the Agence Nationale de la Recherche (grant ANR-18-CE46-0002).
\end{ack}

\appendix
\clearpage

\section{Matrix-variate Posterior Distribution Induced by \hvi}
We derive the parameters of the matrix-variate distribution $q(\Wmatr)=\mathcal{MN}(\Mmatr, \Umatr, \Vmatr)$ of the weight matrix $ \tilde\Wmatr \in \R^{D \times D}$ given by \hvi,
\begin{align}
    \tilde\Wmatr = \Smatr_1 \Hmatr \diag(\tilde\gvect) \Hmatr\Smatr_2 \quad \mathrm{with} \quad \tilde\gvect \sim  \N(\muvect, \Sigmamatr).
\end{align}
The mean $\Mmatr = \Smatr_1\Hmatr\diag(\muvect) \Hmatr\Smatr_2$ derives from the linearity of the expectation.
The covariance matrices $\Umatr$ and $\Vmatr$  are non-identifiable: for any scale factor $s>0$, we have $\mathcal{MN}(\Mmatr, \Umatr, \Vmatr)$ equals $\mathcal{MN}(\Mmatr, s\Umatr, \frac{1}{s}\Vmatr)$. Therefore, we constrain the parameters such that $\Tr(\Vmatr) = 1$. The covariance matrices verify (see e.g. Section~1 in the supplement of~\cite{Ding2014})
\begin{align*}
    \Umatr & = \Exp\left[(\Wmatr-\Mmatr)(\Wmatr-\Mmatr)^\T\right]                       \\
    \Vmatr & = \frac{1}{\Tr(\Umatr)}\Exp\left[(\Wmatr-\Mmatr)^\T(\Wmatr-\Mmatr)\right].
\end{align*}
The Walsh-Hadamard matrix $H$ is symmetric.
Denoting by $\Sigmamatr^{1/2}$ a root of $\Sigmamatr$ and considering $\varepsilonvect\sim\norm(\zerovect,\Imatr)$, we have
\begin{align}
    \Umatr & = \Exp\left[\Smatr_1 \Hmatr \diag(\Sigmamatr^{1/2} \varepsilonvect) \Hmatr \Smatr_2^2 \Hmatr \diag(\Sigmamatr^{1/2} \varepsilonvect) \Hmatr \Smatr_1\right].
    \label{eq:U}
\end{align}
If we define the matrix $\Tmatr_2 \in \R^{D \times D^2}$ where the $i^\text{th}$ row is the column-wise vectorization of the matrix $(\Sigmamatr^{1/2}_{i,j}(\Hmatr {\Smatr_2})_{i,j'})_{j,j'\le D}$. We have
\begin{align*}
     & (\Tmatr_2\Tmatr_2^\T)_{i,i'}  = \sum_{j,j' =1}^{D} \Sigmamatr^{1/2}_{i,j}\Sigmamatr^{1/2}_{i',j} (\Hmatr \Smatr_2)_{i,j'} (\Hmatr \Smatr_2)_{i',j'}                                                                                     \\
     & = \sum_{j,j',j'' =1}^{D} \Sigmamatr^{1/2}_{i,j} (\Hmatr \Smatr_2)_{i,j'} \Exp[\epsilon_j\epsilon_{j''}] \Sigmamatr^{1/2}_{i',j''}(\Hmatr \Smatr_2)_{i',j'}                                                                              \\
     & = \sum_{j'=1}^{D} \Exp\left[ \left(\sum_{j=1}^{D} \epsilon_j \Sigmamatr^{1/2}_{i,j} (\Hmatr \Smatr_2)_{i,j'}\right)\right. \left.\left(\sum_{j''=1}^{D} \epsilon_{j''} \Sigmamatr^{1/2}_{i',j''}(\Hmatr \Smatr_2)_{i',j'}\right)\right] \\
     & = \Exp\left[\left(\diag(\Sigmamatr^{1/2} \varepsilonvect) \Hmatr \Smatr_2^2 \Hmatr \diag(\Sigmamatr^{1/2} \varepsilonvect) \right)_{i,i'}\right].
\end{align*}
Using \eqref{eq:U}, a root of $\Umatr=\Umatr^{1/2}{\Umatr^{1/2}}^\T$ can be found:
\begin{equation}
    \Umatr^{1/2} = \Smatr_1 \Hmatr \Tmatr_2.
\end{equation}
Similarly for $\Vmatr$, we have
\begin{align}
    \Vmatr^{1/2} =\frac{1}{\sqrt{\Tr(\Umatr)}}\Smatr_2 \Hmatr \Tmatr_1, \nonumber \\
    \text{with }  \Tmatr_1 = \left[\footnotesize\begin{matrix}\vect\left(\Sigmamatr_{1,:}\left(\Hmatr\Smatr_1\right)_{1,:}^\T\right)^\T \\
            \vdots                                                                    \\
            \vect\left(\Sigmamatr_{D,:}\left(\Hmatr\Smatr_1\right)_{d,:}^\T\right)^\T
        \end{matrix}\right].
\end{align}

\section{Geometric Interpretation of \hvi}
\label{sec:geometrical-interpr}
The matrix $\Amatr$ in Section 2.2 expresses the linear relationship  between the weights $\Wmatr=\Smatr_1 \Hmatr \Gmatr \Hmatr \Smatr_2$ and the variational random vector $\gvect$, i.e. $\vect(\Wmatr) = \Amatr \gvect$.
Recall the definition of
\begin{equation}
    \Amatr =
    \left[\begin{matrix}
            \Smatr_1\Hmatr\diag(\vvect_1) \\
            \vdots                        \\
            \Smatr_1\Hmatr\diag(\vvect_D)
        \end{matrix}
        \right], ~\text{ with }\vvect_i = (\Smatr_2)_{i,i}(\Hmatr)_{:,i}.
\end{equation}
We show that a $\Lmatr\Qmatr$-decomposition of $\Amatr$ can be explicitly formulated.
\paragraph{Proposition.}
Let $A$ be a $D^2\times D$ matrix such that $\vect(\Wmatr) = \Amatr \gvect$, where $\Wmatr$ is given by $\Wmatr=\Smatr_1\Hmatr\diag(\gvect) \Hmatr\Smatr_2$. Then a $\Lmatr\Qmatr$-decomposition of $\Amatr$ can be formulated as
\begin{align}
    \vect(\Wmatr) & = [s^{(2)}_{i} \Smatr_1 \Hmatr \diag(\hvect_i)]_{i=1,\ldots,D} ~ \gvect\nonumber \\
                  & =  \Lmatr \Qmatr \gvect,
    \label{eq:LQ}
\end{align}
where $\hvect_i$ is the $i^\text{th}$ column of $\Hmatr$,  $\Lmatr = \diag((s^{(2)}_{i} \svect)_{i=1,\ldots,D})$, $\diag(\svect^{(1)}) = \Smatr_1$, $\diag(\svect^{(2)}) = \Smatr_2$, and $\Qmatr = [\Hmatr \diag(\hvect_i)]_{i=1,\ldots,D}$.

\begin{minipage}{.55\textwidth}
\paragraph{Proof.}\textit{
    Equation~\eqref{eq:LQ} derives directly from block matrix and vector operations. As $\Lmatr$ is clearly lower triangular (even diagonal), let us proof that $\Qmatr$ has orthogonal columns. Defining the $d\times d$ matrix $\Qmatr^{(i)} = \Hmatr \diag(\hvect_i)$, we have:
    \begin{align*}
        \Qmatr^\T \Qmatr & = \sum_{i=1}^{D} {\Qmatr^{(i)}}^\T\Qmatr^{(i)}                      \\
                         & = \sum_{i=1}^{D} \diag(\hvect_i){\Hmatr}^\T \Hmatr  \diag(\hvect_i) \\
                         & = \sum_{i=1}^{D} \diag(\hvect_i^2)= \sum_{i=1}^{D} \frac{1}{D}I= I. \\
    \end{align*}
}
\end{minipage}\begin{minipage}{.45\textwidth}\centering
    \includegraphics{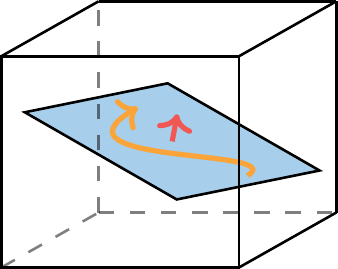}
    \captionof{figure}{Diagrammatic representation of \hvi. The cube represent the high dimensional parameter space. The variational posterior (mean in orange) evolves during optimization in the (blue) subspace whose orientation (red) is controlled by $\Smatr_1$ and $\Smatr_2$.}
    \label{fig:geometry}
\end{minipage}

This decomposition gives direct insight on the role of the Walsh-Hadamard transforms: with complexity $D\log(D)$, they perform fast rotations $\Qmatr$ of vectors living in a space of dimension $D$ (the plane in \cref{fig:geometry}) into a space of dimension $D^2$ (the cube in Figure~\ref{fig:geometry}).
Treated as  parameters  gathered in $\Lmatr$, $\Smatr_1$ and $\Smatr_2$ control the orientation of the subspace by distortion of the canonical axes.

We empirically evaluate the minimum \rmse, as a proxy for some measure of average distance, between $\Wmatr$ and any given point $\Gammamatr$. More precisely, we compute for $\Gammamatr \in \R^{D\times D}$,
\begin{equation}
    \min_{\svect_1,\svect_2,\gvect \in \R^D} {\displaystyle\frac{1}{D}||\Gammamatr -  \diag(\svect_1)\Hmatr\diag(\gvect)\Hmatr\diag(\svect_2)||_\text{Frob}}.
\end{equation}

\begin{figure}[H]
    \centering
    \scriptsize
    \includegraphics{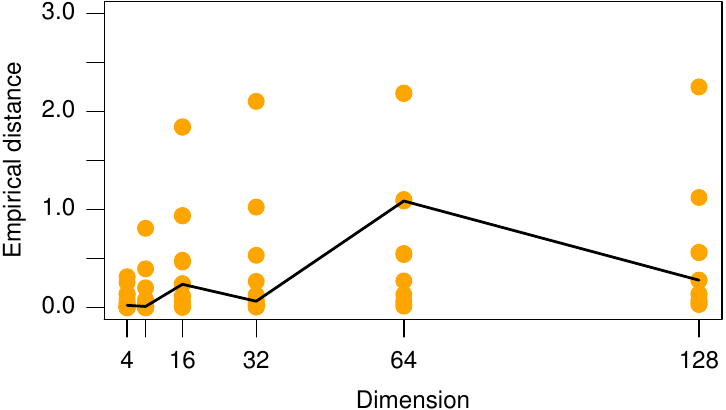}
    \caption{Distribution of the minimum \rmse between $\Smatr_1 \Hmatr \Gmatr \Hmatr \Smatr_2$ and a sample matrix with i.i.d. $\mathcal{U}(-1,1)$ entries. For each dimension, the orange dots represent 20 repetitions. The median distance is displayed in black.  Few outliers (with distance greater than 3.0) appeared, possibly due to imperfect numerical optimization. They were kept for the calculation of the median but not displayed.
        }
    \label{fig:distance}
\end{figure}

\cref{fig:distance} shows this quantity evaluated for $\Gammamatr$ sampled with i.i.d $\mathcal{U}(-1,1)$ with increasing value of $D$.
The bounded behavior suggests that \hvi can approximate any given matrices with a precision that does not increase with the dimension.

\section{Additional Details on Normalizing Flows}
In the general setting, given a probabilistic model with observations $\xvect$, latent variables $\zvect$ and model parameters $\thetavect$, by introducing an approximate posterior distribution $q_\phi(\zvect)$ with parameters $\phi$, the variational lower bound to the log-marginal likelihood is defined as
\begin{align}
    \KL\{q_\phi(\zvect)& || p(\zvect|\xvect)\}  = \Exp_{q_{\phi}(\zvect)} \left[\log q_{\phi}(\zvect) - \log p(\zvect|\xvect) \right] \nonumber                                \\
                                              & = \Exp_{q_{\phi}(\zvect)} \left[\log q_{\phi}(\zvect) -  \log p_{\thetavect}(\xvect,\zvect) - \log p(\xvect) \right] \nonumber \\
                                              & \leq -\Exp_{q_{\phi}(\zvect)} \left[\log p_{\thetavect}(\xvect|\zvect) - \log q_{\phi}(\zvect) + \log p(\zvect)\right]
    \end{align}
where $p_{\thetavect}(\xvect|\zvect)$ is the likelihood function with $\thetavect$ model parameters and $p(\zvect)$ is the prior on the latents.
The objective is then to minimize the negative variational bound (\nelbo):
\begin{align}
    \mathcal{L}(\thetavect, \phi) = -\Exp_{q_{\phi}(\zvect)} \log p_{\thetavect}(\xvect|\zvect) + \KL\{q_{\phi}\left (\zvect) || p(\zvect))\right \}\,.
\end{align}

Consider an invertible, continuous and differentiable function $f:\R^D\rightarrow\R^D$.
Given $\tilde\zvect_0 \sim q(\zvect_0)$, then $\tilde\zvect_1 = f(\tilde\zvect_0)$ follows $q(\zvect_1)$ defined as
\begin{align}
    q(\zvect_1) = q(\zvect_0) \left| \det\frac{\partial f}{\partial \zvect_0} \right|^{-1}\,.
\end{align}
As a consequence, after $K$ transformations the log-density of the final distribution is
\begin{align}
    \log q(\zvect_K) = \log q(\zvect_0) - \sum_{k=1}^K \log \left| \det \frac{\partial f_{k-1}}{\partial \zvect_{k-1}} \right|\,.
\end{align}
We shall define $f_k(\zvect_{k-1}; \lambdavect_k)$ the $k^{\mathrm{th}}$ transformation which takes input from the previous flow $\zvect_{k-1}$ and has parameters $\lambdavect_k$.
The final variational objective is
\begingroup
\allowdisplaybreaks
\begin{align}
    &\mathcal{L}(\thetavect, \phi) = -\Exp_{q_{\phi}(\zvect)}[ \log p_{\thetavect}(\xvect|\zvect) ] + \KL\{q_{\phi}\left (\zvect) || p(\zvect)\}\right ) \nonumber                             \\
                                  & = \Exp_{q_{\phi}(\zvect|\xvect)} [ - \log p_{\thetavect}(\xvect | \zvect) - \log p(\zvect) + \log q_\phi(\zvect )  ] \nonumber                             \\
                                  & = \Exp_{q_0(\zvect_0)} [ - \log p_{\thetavect}(\xvect|\zvect_K) - \log p(\zvect_K) + \log q_K(\zvect_K)]  \nonumber                                        \\
                                  & = \Exp_{q_0(\zvect_0)} \left[ - \log p_{\thetavect}(\xvect|\zvect_K) - \log p(\zvect_K) + \log q_0(\zvect_0)\right. \nonumber                                \\
                                  & ~~~~~~~~~~~~~~~~~~~~~~ -   \left.\sum_{k=1}^K\log \left|\det \frac{\partial f_k(\zvect_{k-1};\lambdavect_k)}{\partial \zvect_{k-1}} \right|\right] \nonumber \\
                                  & = - \Exp_{q_0(\zvect_0)}  \log p_{\thetavect}(\xvect|\zvect) + \KL\{q_0(\zvect_0)||p(\zvect_K)\}  \nonumber \\
                                  & ~~~~~~~~~~~~~~~~~~~~~~ -  \Exp_{q_0(\zvect_0)} \sum_{k=1}^K \log \left|\det \frac{\partial f_k(\zvect_{k-1};\lambdavect_k)}{\partial \zvect_{k-1}} \right|\,.
\end{align}
\endgroup

Setting the initial distribution $q_0$ to a fully factorized Gaussian $\N(\zvect_0|\muvect, \sigmavect\Ivect)$ and assuming a Gaussian prior on the generated $\zvect_K$, the \KL term is analytically tractable.
A possible family of transformation is the {\em planar flow} \citep{Rezende2015}.
For the {\em planar flow}, $f$ is defined as
\begin{align}
    f(\zvect) = \zvect + \uvect h(\wvect^\T\zvect + b)\,,
\end{align}
where $\lambda = [ \uvect \in \R^D,\, \wvect \in \R^D,\, b \in \R ]$ and $h(\cdot) = \tanh(\cdot)$. This is equivalent to a residual layer with single neuron \textsc{mlp} -- as argued by \citet{Kingma2016}.
The log-determinant of the Jacobian of $f$ is
\begin{align}
    \log \left| \det \frac{\partial f}{\partial \zvect} \right| & = \left| \det (\Ivect  + \uvect[h^\prime(\wvect^\T\zvect + b)\wvect]^\T) \right| \nonumber \\
                                                                & = \left| 1 + \uvect^\T\wvect h^\prime(\wvect^\T\zvect + b)\right|\,.
\end{align}
Although this is a simple flow parameterization, a planar flow requires only $\bigO(D)$ parameters and thus it does not increase the time/space complexity of \hvi.
Alternatives can be found in \citep{Rezende2015, Berg2018, Kingma2016, Louizos2017}.

\section{Additional Results}

\subsection{Experimental Setup for Bayesian DNN}
\begin{minipage}{.42\textwidth}
    The experiments on Bayesian \dnn are run with the following setup.
    For \hvi, we used a zero-mean prior over $\gvect$ with fully factorized covariance $\lambda \Imatr$; $\lambda = 10^{-5}$ was chosen to obtain sensible variances in the output layer.
    It is possible to design a prior over $\gvect$ such that the prior on $\Wmatr$ has constant marginal variance and low correlations although empirical evaluations showed not to yield a significant improvement compared to the previous (simpler) choice.
    In the final implementation of \hvi that we used in all experiments, $\Smatr_1$ and $\Smatr_2$ are optimized.
    The dropout rate of \mcd is set to 0.005.
    We used classic Gaussian likelihood with optimized noise variance for regression and softmax likelihood for classification.
\end{minipage}\hfill
\begin{minipage}{.54\textwidth}
    \sc
    \scriptsize
    \centering
    \captionof{table}{List of dataset used in the experiments}
    \label{tab:datasets}
    \vspace{0pt}
    \begin{tabular}{r|lrrr}
    \toprule
    \textbf{ name } & \textbf{ task } & \textbf{ n. } & \textbf{ d-in } & \textbf{ d-out } \\
    \midrule
    eeg             & Class.          & 14980         & 14              & 2                \\
    magic           & Class.          & 19020         & 10              & 2                \\
    miniboo         & Class.          & 130064        & 50              & 2                \\
    letter          & Class.          & 20000         & 16              & 26               \\
    drive           & Class.          & 58509         & 48              & 11               \\
    mocap           & Class.          & 78095         & 37              & 5                \\
    cifar10         & Class.          & 60000         & 3 $\times$ 28 $\times$ 28 & 10      \\
    \midrule
    boston          & Regr.           & 506           & 13              & 1                \\
    concrete        & Regr.           & 1030          & 8               & 1                \\
    energy          & Regr.           & 768           & 8               & 2                \\
    kin8nm          & Regr.           & 8192          & 8               & 1                \\
    naval           & Regr.           & 11934         & 16              & 2                \\
    powerplant      & Regr.           & 9568          & 4               & 1                \\
    protein         & Regr.           & 45730         & 9               & 1                \\
    yacht           & Regr.           & 308           & 6               & 1                \\
    \midrule
    borehol         & Regr.           & 200000        & 8               & 1                \\
    hartman6        & Regr.           & 30000         & 6               & 1                \\
    rastrigin5      & Regr.           & 10000         & 5               & 1                \\
    robot           & Regr.           & 150000        & 8               & 1                \\
    otlcircuit      & Regr.           & 20000         & 6               & 1                \\
    \bottomrule
\end{tabular}
    \end{minipage}

    Training is performed for 500 steps with fixed noise variance and for other 50000 steps with optimized noise variance.
Batch size is fixed to 64 and for the estimation of the expected loglikelihood we used 1 Monte Carlo sample at train-time and 64 Monte Carlo samples at test-time.
We choose the Adam optimizer \citep{Kingma2015b} with exponential learning rate decay $\lambda_{t+1} = \lambda_0(1 + \gamma t) ^{-p}$, with $\lambda_0=0.001$, $p=0.3$, $\gamma=0.0005$ and $t$ being the current iteration.

Similar setup was also used for the Bayesian \cnn experiment.
The only differences are the batch size -- increased to 256 -- and the optimizer, which is run without learning rate decay.

\subsection{Regression Experiments on Shallow Models}
For a complete experimental evaluation of \hvi, we also use the experimental setup proposed by \citet{Lobato2015} and adopted in several other works \citep{Gal16, Louizos2016, Zhang2018}.
In this configuration, we use one hidden layer with 50 hidden units for all datasets with the exception of \name{protein} where the number of units is increased to 100.
Results are reported in \autoref{tab:lobato-setup}.

\begin{table*}[ht]
    \sc\scriptsize
    \caption{Test \rmse and test \mnll for regression datasets following the setup in \citep{Lobato2015}.}
    \label{tab:lobato-setup}
    \renewcommand{\tabcolsep}{1ex}
\begin{tabular}{l||rrrr|rrrr}
    \toprule
    {}         & \multicolumn{4}{r|}{{test error}} & \multicolumn{4}{r}{{test mnll}}                                                                                                                                                                              \\
    model      & mcd                               & mfg                             & nng                            & whvi                            & mcd                             & mfg                             & nng              & whvi             \\
    dataset    &                                   &                                 &                                &                                 &                                 &                                                                       \\
    \midrule
    boston     & $3.40$ \scalebox{.8}{$(0.66)$}    & $3.04$ \scalebox{.8}{$(0.64)$}  & $2.74$ \scalebox{.8}{$(0.12)$} & $2.56$  \scalebox{.8}{$(0.15)$} & $ 5.04$ \scalebox{.8}{$(1.76)$} & $ 3.19$ \scalebox{.8}{$(0.89)$} & $ 2.45$ \scalebox{.8}{$(0.03)$} & $ 2.55$ \scalebox{.8}{$(0.15)$} \\
    concrete   & $4.60$ \scalebox{.8}{$(0.53)$}    & $5.24$ \scalebox{.8}{$(0.53)$}  & $5.02$ \scalebox{.8}{$(0.12)$} & $5.01$  \scalebox{.8}{$(0.25)$} & $ 2.96$ \scalebox{.8}{$(0.23)$} & $ 3.03$ \scalebox{.8}{$(0.15)$} & $ 3.04$ \scalebox{.8}{$(0.02)$} & $ 2.95$ \scalebox{.8}{$(0.06)$} \\
    energy     & $1.18$ \scalebox{.8}{$(0.03)$}    & $1.52$ \scalebox{.8}{$(0.09)$}  & $0.48$ \scalebox{.8}{$(0.02)$} & $1.20$  \scalebox{.8}{$(0.07)$} & $ 3.00$ \scalebox{.8}{$(0.07)$} & $ 3.49$ \scalebox{.8}{$(0.11)$} & $ 1.42$ \scalebox{.8}{$(0.00)$} & $ 3.01$ \scalebox{.8}{$(0.12)$} \\
    kin8nm     & $0.09$ \scalebox{.8}{$(0.00)$}    & $0.10$ \scalebox{.8}{$(0.00)$}  & $0.08$ \scalebox{.8}{$(0.00)$} & $0.12$  \scalebox{.8}{$(0.01)$} & $-1.09$ \scalebox{.8}{$(0.04)$} & $-1.01$ \scalebox{.8}{$(0.04)$} & $-1.15$ \scalebox{.8}{$(0.00)$} & $-0.78$ \scalebox{.8}{$(0.10)$} \\
    naval      & $0.00$ \scalebox{.8}{$(0.00)$}    & $0.01$ \scalebox{.8}{$(0.00)$}  & $0.00$ \scalebox{.8}{$(0.00)$} & $0.01$  \scalebox{.8}{$(0.00)$} & $-9.93$ \scalebox{.8}{$(0.01)$} & $-6.48$ \scalebox{.8}{$(0.02)$} & $-7.08$ \scalebox{.8}{$(0.03)$} & $-6.25$ \scalebox{.8}{$(0.01)$} \\
    powerplant & $4.20$ \scalebox{.8}{$(0.12)$}    & $4.23$ \scalebox{.8}{$(0.13)$}  & $3.89$ \scalebox{.8}{$(0.04)$} & $4.11$  \scalebox{.8}{$(0.12)$} & $ 2.76$ \scalebox{.8}{$(0.03)$} & $ 2.77$ \scalebox{.8}{$(0.03)$} & $ 2.78$ \scalebox{.8}{$(0.01)$} & $ 2.74$ \scalebox{.8}{$(0.03)$} \\
    protein    & $4.35$ \scalebox{.8}{$(0.04)$}    & $4.74$ \scalebox{.8}{$(0.05)$}  & $4.10$ \scalebox{.8}{$(0.00)$} & $4.64$  \scalebox{.8}{$(0.07)$} & $ 2.80$ \scalebox{.8}{$(0.01)$} & $ 2.89$ \scalebox{.8}{$(0.01)$} & $ 2.84$ \scalebox{.8}{$(0.00)$} & $ 2.86$ \scalebox{.8}{$(0.01)$} \\
    yacht      & $1.72$ \scalebox{.8}{$(0.32)$}    & $1.78$ \scalebox{.8}{$(0.45)$}  & $0.98$ \scalebox{.8}{$(0.08)$} & $0.96$  \scalebox{.8}{$(0.20)$} & $ 2.73$ \scalebox{.8}{$(0.74)$} & $ 2.02$ \scalebox{.8}{$(0.46)$} & $ 2.32$ \scalebox{.8}{$(0.00)$} & $ 1.28$ \scalebox{.8}{$(0.22)$} \\
    \bottomrule
\end{tabular}

\end{table*}
\clearpage

\subsection{ConvNets architectures}

\begin{figure*}[ht!]
    \centering
    \resizebox{\textwidth}{!}{\includegraphics{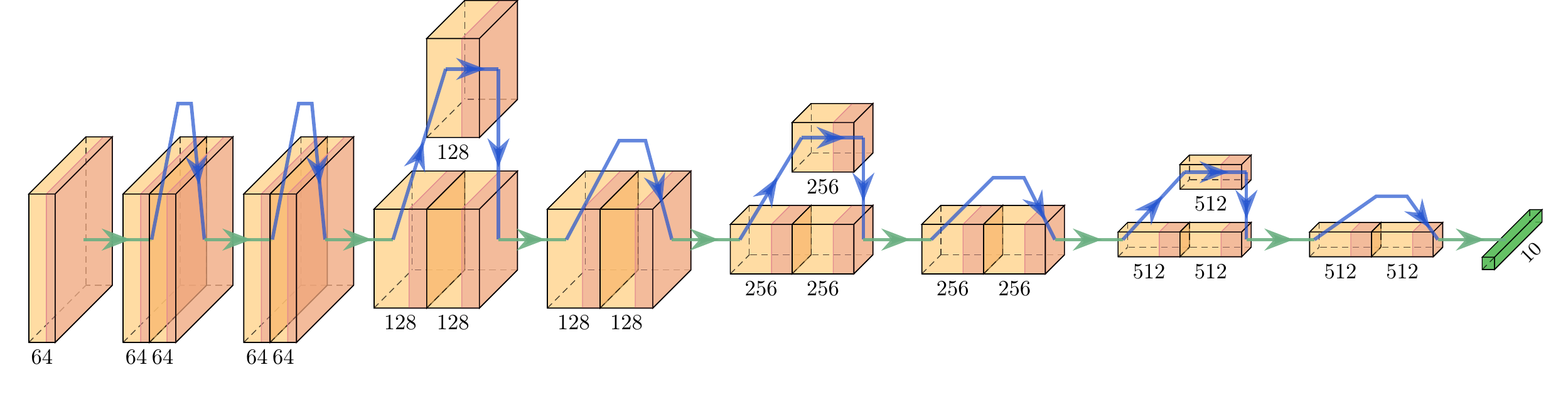}
    }
    \caption{
        Architecture layout of \resnet18.
    }
    \label{fig:alexnet-architecture}
\end{figure*}

For the experiments on Bayesian convolutional neural networks, we used architectures adapted to \cifart (see Tables~\ref{tb:alexnet}, \ref{tb:vgg} and~\ref{tb:resnet}).

\begin{figure*}[ht!]
\begin{minipage}[t]{.32\textwidth}
    \scriptsize
    \centering
    \sc
    \captionof{table}[]{\alexnet}
    \label{tb:alexnet}
    \begin{tabular}{rc}
        \toprule
        \textbf{layer} & \textbf{dimensions}                \\
        \midrule
        conv           & $64 \times 3 \times 3 \times 3$    \\
        maxpool        &                                    \\
        \midrule
        conv           & $192 \times 64 \times 3 \times 3$  \\
        maxpool        &                                    \\
        \midrule
        conv           & $384 \times 192 \times 3 \times 3$ \\
        conv           & $256 \times 384 \times 3 \times 3$ \\
        conv           & $256 \times 256 \times 3 \times 3$ \\
        maxpool        &                                    \\
        \midrule
        linear         & $4096 \times 4096$                 \\
        linear         & $4096 \times 4096$                 \\
        linear         & $10 \times 4096$                   \\
        \bottomrule
    \end{tabular}
\end{minipage}
\begin{minipage}[t]{.32\textwidth}
    \scriptsize
    \centering
    \sc
    \captionof{table}[]{\vgg}
    \label{tb:vgg}
    \begin{tabular}{rc}
        \toprule
        \textbf{layer} & \textbf{dimensions}               \\
        \midrule
        conv           & $32 \times 3 \times 3 \times 3$   \\
        conv           & $32 \times 32 \times 3 \times 3$  \\
        maxpool        &                                   \\
        \midrule
        conv           & $64 \times 32 \times 3 \times 3$   \\
        conv           & $64 \times 64 \times 3 \times 3$  \\
        maxpool        &                                   \\
        \midrule
        conv           & $128 \times 64 \times 3 \times 3$  \\
        conv           & $128 \times 128 \times 3 \times 3$ \\
        conv           & $128 \times 128 \times 3 \times 3$ \\
        maxpool        &                                   \\
        \midrule
        conv           & $256 \times 128 \times 3 \times 3$  \\
        conv           & $256 \times 256 \times 3 \times 3$ \\
        conv           & $256 \times 256 \times 3 \times 3$ \\
        maxpool        &                                   \\
        \midrule
        conv           & $256 \times 256 \times 3 \times 3$  \\
        conv           & $256 \times 256 \times 3 \times 3$ \\
        conv           & $256 \times 256 \times 3 \times 3$ \\
        maxpool        &                                   \\
        \midrule
        linear         & $10 \times 256$                   \\
        \bottomrule
    \end{tabular}
\end{minipage}
\begin{minipage}[t]{.32\textwidth}
    \centering
    \scriptsize
    \sc
    \captionof{table}[]{\resnet~18}
    \label{tb:resnet}
    \begin{tabular}{rc}
        \toprule
        \textbf{layer} & \textbf{dimensions} \\
        \midrule
        resnet block   &
        $\begin{bmatrix}
                3 \times 3, 64 \\
                3 \times 3, 64 \\
            \end{bmatrix} \times 2 $\\
        \midrule
        resnet block   &
        $\begin{bmatrix}
                3 \times 3, 128 \\
                3 \times 3, 128 \\
            \end{bmatrix} \times 2 $\\
        \midrule
        resnet block   &
        $\begin{bmatrix}
                3 \times 3, 256 \\
                3 \times 3, 256 \\
            \end{bmatrix} \times 2 $\\
        \midrule
        resnet block   &
        $\begin{bmatrix}
                3 \times 3, 512 \\
                3 \times 3, 512 \\
            \end{bmatrix} \times 2 $\\
        \midrule
        avgpool & \\
        linear         & $10 \times 512$     \\
        \bottomrule
    \end{tabular}
\end{minipage}
\end{figure*}

\subsection{Results - Gaussian Processes with Random Feature Expansion}

\begin{table}[t]
    \caption{Complexity table for \gps with random feature and inducing points approximations. In the case of random features, we include both the complexity of computing random features and the complexity of treating the linear combination of the weights variationally (using \vi and \hvi).}
    \label{tab:complexityGP}
    \centering
    {
        \sc
        \scriptsize
        \begin{tabular}{lcc}
            \toprule
                                     & \multicolumn{2}{c}{Complexity}                                                                     \\
            {}                       & Space                                               & Time                                         \\
            \midrule
            Mean field \xspace- rf  & $\bigO(D_\text{IN}N_\text{RF}) + \bigO(N_\text{RF}D_\text{OUT})$        & $\bigO(D_\text{IN}N_\text{RF}) + \bigO(N_\text{RF}D_\text{OUT})$ \\
            \hvi \xspace- rf         & $\bigO(D_\text{IN}N_\text{RF}) + \bigO(\sqrt{N_\text{RF}}D_\text{OUT})$ & $\bigO(D_\text{IN}N_\text{RF}) + \bigO(D_\text{OUT}\log{{N_\text{RF}}})$                     \\
            Inducing points          & $\bigO(M)$                                          & $\bigO(M^3)$                                 \\
            \bottomrule
        \end{tabular}
        \label{tab:complexityGP}
    }
    \vspace{1ex}\par
    \scriptsize\normalfont
    Note: $M$ is the number of pseudo-data/inducing points and $N_{RF}$ is the number of random features used in the kernel approximation.
\end{table}

We test \hvi for scalable \gp inference,
by focusing on \gps with random feature expansions \citep{LazaroGredilla10}.
In \gp models, latent variables $\fvect$ are given a prior $p(\fvect) = \norm(\zerovect | \Kmatr)$; the assumption of zero mean can be easily relaxed.
Given a random feature expansion of the kernel martix, say $\Kmatr \approx \Phimatr \Phimatr^{\T}$, the latent variables can be rewritten as:
\begin{equation}
    \fvect = \Phivect \wvect
\end{equation}
with $\wvect\sim \norm(\zerovect,\Imatr)$. The random features $\Phivect$ are constructed by randomly projecting the input matrix $\Xmatr$ using a Gaussian random matrix $\Omegamatr$ and applying a nonlinear transformation, which depends on the choice of the kernel function.
The resulting model is now linear, and considering regression problems such that $\yvect = \fvect +\bm{\varepsilon}$ with $\bm{\varepsilon} \sim \norm(\zerovect, \sigma^2 \Imatr)$, solving \gps for regression becomes equivalent to solving standard linear regression problems.
For a given set of random features, we treat the weights of the resulting linear layer variationally and evaluate the performance of \hvi.

\begin{figure}[b!]
    \tiny
    \centering
    \setlength\figureheight{.27\textwidth}
    \setlength\figurewidth{.26\textwidth}
    \pgfplotsset{every axis title/.append style={yshift=-1ex}}
    \subfigure{
        \includegraphics{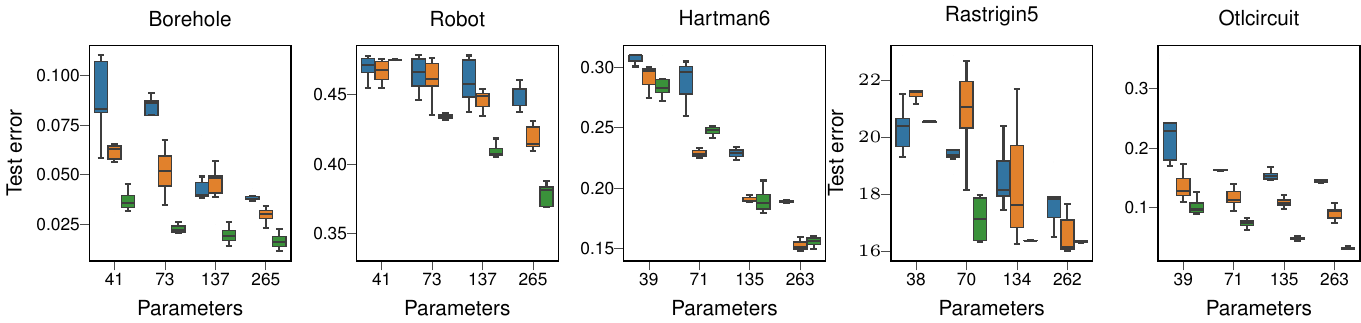}
    }\vspace{-2mm}
    \subfigure{
        \includegraphics{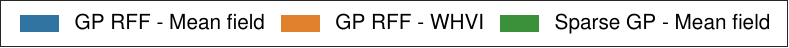}
    }

    \subfigure{
        \includegraphics{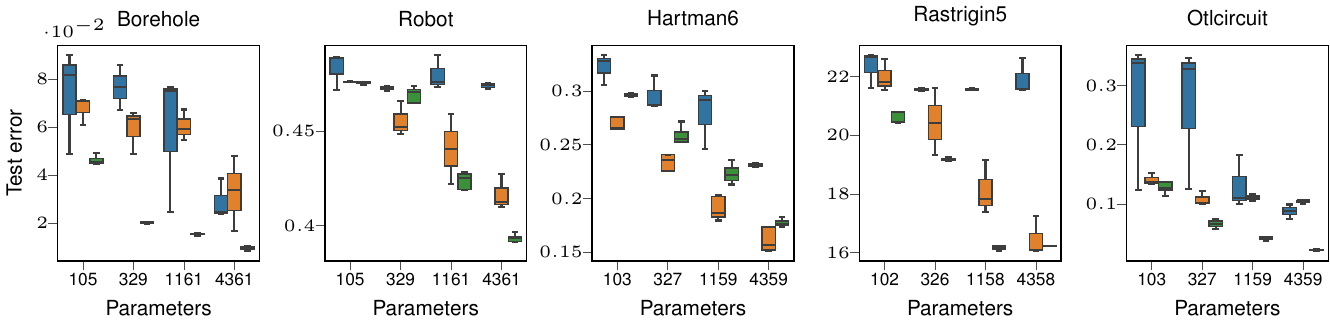}
    }\vspace{-2mm}
    \subfigure{
        \includegraphics{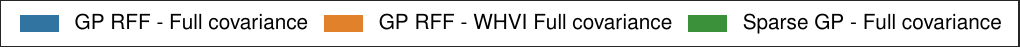}
    }
    \caption{Comparison of test error w.r.t. the number model parameters (\textit{top:} mean field, \textit{bottom:} full covariance).}
    \label{fig:vfastgp_wrt_nparam}
\end{figure}
By reshaping the vector of parameters $\wvect$ of the linear model into a $D \times D$ matrix, \hvi allows for the linearized \gp model to reduce the number of parameters to optimize (see \cref{tab:complexityGP}).
We compare \hvi with two alternatives; one is \vi of the Fourier features \gp expansion that uses less random features to match the number of parameters used in \hvi, and another is the sparse Gaussian process implementation of \name{gpflow} \citep{GPflow2017} with a number of inducing points (rounded up) to match the number of parameters used in \hvi.

We report the results on five datasets ($10000\leq N\leq 200000$, $5\leq D\leq 8$, see \cref{tab:datasets}).
The data sets are generated from space-filling evaluations of well known functions in analysis of computer experiments (see e.g. \cite{simulationlib}). Dataset splitting in training and testing points is  random uniform, 20\% versus 80 \%. The input variables are rescaled between 0 and 1. The output values are standardized for training. All \gps have the same prior (centered \gp  with \rbf covariance), initialized with equal hyperparameter values: each of the $D$ lengthscale to $\sqrt{D/2}$, the \gp variance to $1$, the Gaussian likelihood standard deviation to $0.02$ (prior observation noise). The training is performed with 12000 steps of Adam optimizer. The observation noise is fixed for the first 10000 steps. Learning rate is $6\times 10^{-4}$, except for the dataset {\sc{hartman6}} with a learning rate of $5\times 10^{-3}$. Sparse \gps  are run with whitened representation of the inducing points.

The results are shown in \cref{fig:vfastgp_wrt_nparam} with diagonal covariance for the three variational posteriors and with full covariance.
In both mean field and full covariance settings, this variant of \hvi using the reshaping of $\Wmatr$ into a column largely outperforms the direct \vi of Fourier features.
However, it appears that this improvement of the random feature inference for \gps is not enough to reach the performance of \vi using inducing points.
Inducing point approximations are based on the Nystro\"om approximation of kernel matrices, which are known to lead to lower approximation error on the elements on the kernel matrix compared to random features approximations.
This is the reason we attribute to the lower performance of \hvi compared to inducing points approximations in this experiment.

\subsection{Extended results - \dnns}
\begin{figure*}[b]
    \tiny
    \tikzset{mark size=1}
    \pgfplotsset{every axis title/.append style={yshift=-2ex}}
    \centering
    \setlength\figureheight{.20\textwidth}
    \setlength\figurewidth{.28\textwidth}
    \subfigure{
        \includegraphics{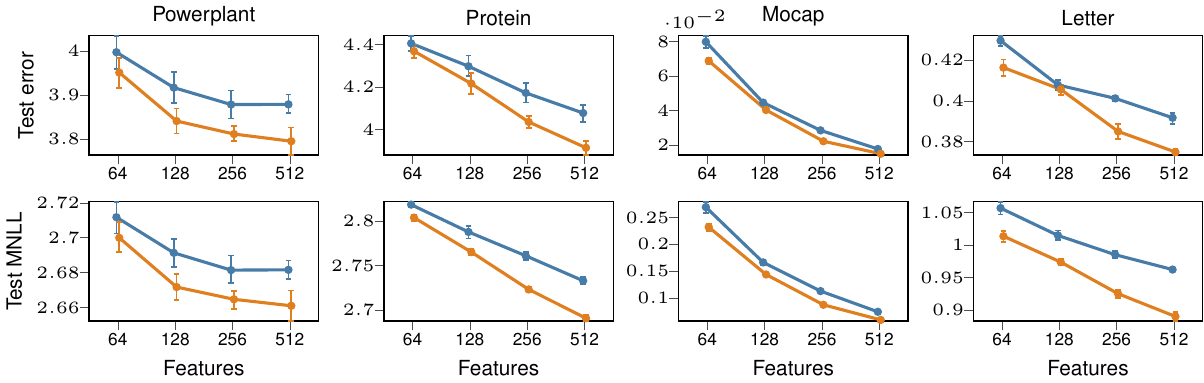}
    }\vspace{-3ex}
    \subfigure{
        \includegraphics{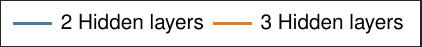}
    }
    \caption{Analysis of model capacity for different features and hidden layers.
    }
    \label{fig:exp-model-capacity}
\end{figure*}

\begin{figure*}
    \tiny
    \tikzset{mark size=1}
    \pgfplotsset{every axis title/.append style={yshift=-2ex}}
    \pgfplotsset{every x tick label/.append style={font=\fontsize{2}{4}\selectfont}}
    \pgfplotsset{every y tick label/.append style={font=\fontsize{2}{4}\selectfont}}
    \centering
    \setlength\figureheight{.225\textwidth}
    \setlength\figurewidth{.505\textwidth}
    \subfigure{
        \includegraphics{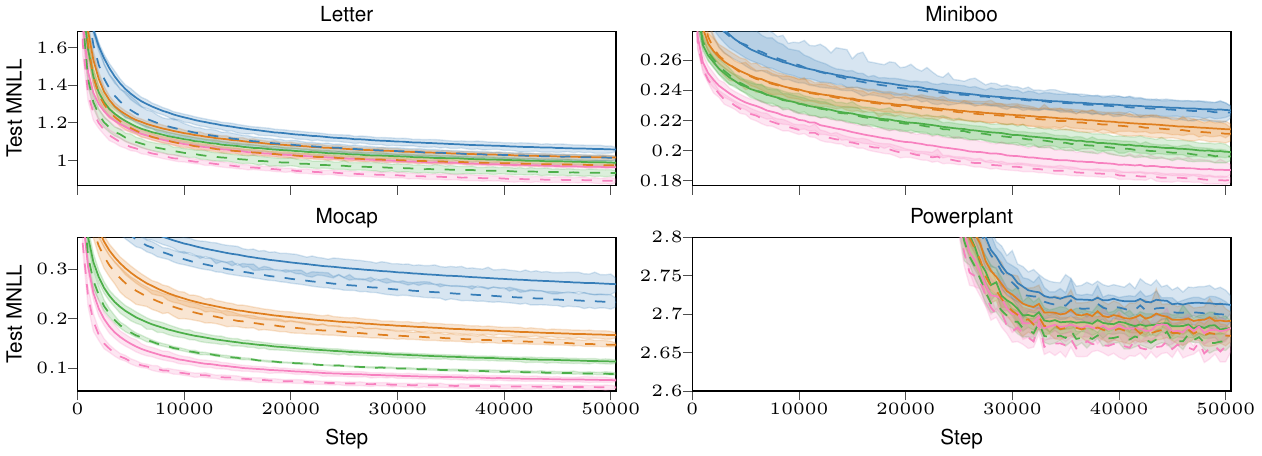}
    }\vspace{-3mm}
    \subfigure{
        \includegraphics{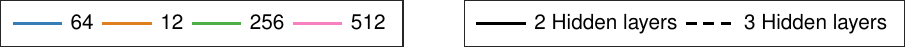}
    }

    \caption{Comparison of test performance. Being able to increase features and hidden layers without worrying about overfitting/overparametrize the model is advantageous not only at convergence but during the entire learning procedure}
    \label{fig:selection_test_mnll_curves}
\end{figure*}

Being able to increase width and depth of a model without drastically increasing the number of variational parameters is one of the competitive advantages of \hvi. 
\cref{fig:exp-model-capacity} shows the behavior of \hvi for different network configurations. At test time, increasing the number of hidden layers and the numbers of hidden features allow the model to avoid overfitting while delivering better performance.
This evidence is also supported by the analysis of the test \mnll during optimization of the \elbo, as showed in \cref{fig:selection_test_mnll_curves}.

Thanks to \hvi structure of the weights matrices, expanding and deepening the model is beneficial not only at convergence but during the entire learning procedure as well.
Furthermore, the derived \nelbo is still a valid lower bound of the true marginal likelihood and, therefore, a suitable objective function for model selection.
Differently from the issue addressed in \citep{Rossi2018}, during our experiments we didn't experience problems regarding initialization of variational parameters. 
We claim that this is possible thanks to both the reduced number of parameters and the effect of the Walsh-Hadamard transform.

\setlength\figureheight{.22\textwidth}
\setlength\figurewidth{.28\textwidth}
\begin{figure}
    \tiny
    \centering
    \pgfplotsset{every axis title/.append style={yshift=-2ex}}
    \includegraphics{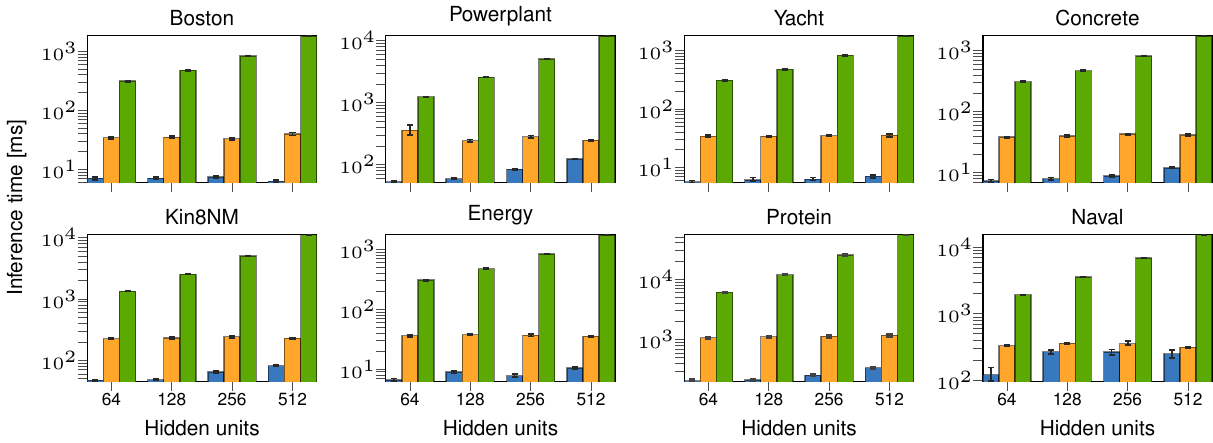}
    \includegraphics{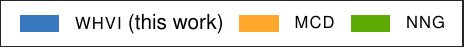}
    \captionof{figure}{Inference time on the test set with 128 batch size and 64 Monte Carlo samples. Experiment repeated 100 times. Additional datasets available in the Supplement.
        }
    \label{fig:inference_time}
\end{figure}

\begin{figure}[h!]
    \centering\tiny
    \pgfplotsset{every axis title/.append style={yshift=-1ex}}
    \setlength\figureheight{.33\textwidth}
    \setlength\figurewidth{.4\textwidth}
    \subfigure{
    \includegraphics{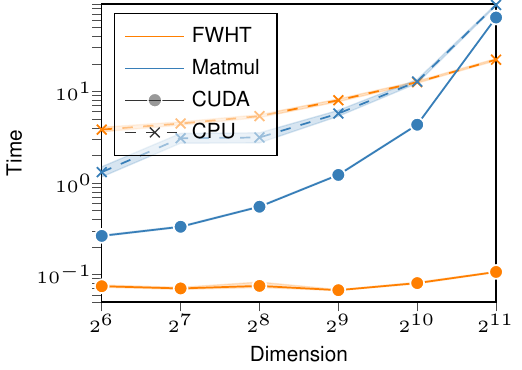}
    }\hspace{-.5cm}
    \subfigure{
    \setlength\figurewidth{.5\textwidth}
    \setlength\figureheight{.2\textwidth}
    \includegraphics{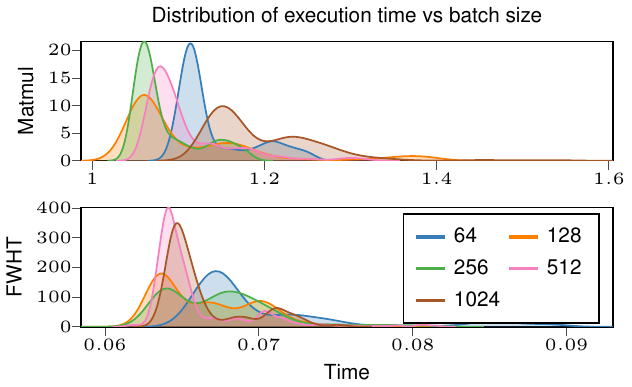}
    } 
    \caption{On the (\textbf{left}), time performance versus number of features (\textsc{d}) with batch size fixed to 512. On the (\textbf{right}) distribution of inference time versus batch size (\name{d}=512) with \name{matmul} and \name{fwht} on \name{gpu}.}
    \label{fig:timing-fwht}
\end{figure}

\paragraph{Timing profiling of the Fast Walsh-Hadamard transform}
Key to the log-linear time complexity is the Fast Walsh-Hadamard transform, which allows to perform the operation $\Hmatr\xvect$ in $\bigO(D\log D)$ time without requiring to generate and store $\Hmatr$.
For our experimental evaluation, we implemented a \name{fwht} operation in \pytorch (v. \texttt{0.4.1}) in C++ and CUDA to leverage the full computational capabilities of modern \textsc{gpu}s. 
\cref{fig:timing-fwht} presents a timing profiling of our implementation versus the naive \texttt{matmul} (batch size of 512 samples and profiling repeated 1000 times). 
The breakeven point for the \name{cpu} implementation is in the neighborhood of $512$/$1024$ features, while on \name{gpu} we see \name{fwht} is consistently faster.
 
\begin{landscape}
    \begin{table}
        \sc
        \tiny
        \centering
        \setlength\tabcolsep{3ex}
        \caption{Test error of Bayesian \dnn with 2 hidden layers on regression datasets. \textsc{nf}: number of hidden features}
        \begin{tabular}{ll||rrrrrrrr}
\toprule
      & {} & \multicolumn{8}{r}{test error} \\
      & dataset &         boston &        concrete &         energy &         kin8nm &          naval &        powerplant &        protein &          yacht \\
model & nf &                &                 &                &                &                &                   &                &                \\
\midrule
mcd & 64  &  $3.80 \pm 0.88$ &   $5.43 \pm 0.69$ &  $2.13 \pm 0.12$ &  $0.17 \pm 0.22$ &  $0.07 \pm 0.00$ &    -- &  $4.36 \pm 0.12$ &   $2.02 \pm 0.51$ \\
      & 128 &  $3.91 \pm 0.86$ &   $5.12 \pm 0.79$ &  $2.07 \pm 0.11$ &  $0.09 \pm 0.00$ &  $0.30 \pm 0.30$ &    $3.97 \pm 0.14$ &  $4.23 \pm 0.10$ &   $1.90 \pm 0.54$ \\
      & 256 &  $3.62 \pm 1.01$ &   $5.03 \pm 0.74$ &  $2.04 \pm 0.11$ &  $0.10 \pm 0.00$ &  $0.07 \pm 0.00$ &  $3.91 \pm 0.11$ &  $4.09 \pm 0.11$ &   $2.09 \pm 0.66$ \\
      & 512 &  $3.56 \pm 0.85$ &   $4.81 \pm 0.79$ &  $2.03 \pm 0.12$ &  $0.09 \pm 0.00$ &  $0.07 \pm 0.00$ &    $\bm{3.90 \pm 0.10}$ &  $\bm{3.87 \pm 0.11}$ &   $2.09 \pm 0.55$ \\
mfg & 64  &  $4.06 \pm 0.72$ &   $6.87 \pm 0.54$ &  $2.42 \pm 0.12$ &  $0.11 \pm 0.00$ &  $0.01 \pm 0.00$ &     $4.38 \pm 0.12$ &  $4.85 \pm 0.12$ &   $4.31 \pm 0.62$ \\
      & 128 &  $4.47 \pm 0.85$ &   $8.01 \pm 0.41$ &  $3.10 \pm 0.14$ &  $0.12 \pm 0.00$ &  $0.01 \pm 0.00$ &     $4.52 \pm 0.13$ &  $4.93 \pm 0.11$ &   $7.01 \pm 1.22$ \\
      & 256 &  $5.27 \pm 0.98$ &   $9.41 \pm 0.54$ &  $4.03 \pm 0.10$ &  $0.13 \pm 0.00$ &  $0.01 \pm 0.00$ &     $4.79 \pm 0.12$ &  $5.07 \pm 0.12$ &   $8.71 \pm 1.31$ \\
      & 512 &  $6.04 \pm 0.90$ &  $10.84 \pm 0.46$ &  $4.90 \pm 0.11$ &  $0.16 \pm 0.00$ &  $0.01 \pm 0.00$ &     $5.53 \pm 0.16$ &  $5.26 \pm 0.10$ &  $10.34 \pm 1.45$ \\
nng & 64  &  $3.20 \pm 0.26$ &   $6.90 \pm 0.59$ &  $1.54 \pm 0.18$ &    ${0.07 \pm 0.00}$ &  $\bm{0.00 \pm 0.00}$ &     ${3.94 \pm 0.05}$ &  $3.90 \pm 0.02$ &   $3.57 \pm 0.70$ \\
      & 128 &  $3.56 \pm 0.43$ &   $8.21 \pm 0.55$ &  $1.96 \pm 0.28$ &  ${0.07 \pm 0.00}$ &  $\bm{0.00 \pm 0.00}$ &     $4.23 \pm 0.09$ &  $4.57 \pm 0.47$ &   $5.16 \pm 1.48$ \\
      & 256 &  $4.87 \pm 0.94$ &   $8.18 \pm 0.57$ &  $3.41 \pm 0.55$ &  ${0.07 \pm 0.00}$ &  $\bm{0.00 \pm 0.00}$ &     $4.07 \pm 0.00$ &  $4.88 \pm 0.00$ &   $5.60 \pm 0.65$ \\
      & 512 &  $5.19 \pm 0.62$ &  $11.67 \pm 2.06$ &  $5.12 \pm 0.37$ &  $0.10 \pm 0.00$ &  $\bm{0.00 \pm 0.00}$ &     $4.97 \pm 0.00$ &            -- &   $5.91 \pm 0.80$ \\
whvi & 64  &  $3.33 \pm 0.82$ &   $5.24 \pm 0.77$ &  $0.73 \pm 0.11$ &  $0.08 \pm 0.00$ &  $0.01 \pm 0.00$ &     $4.07 \pm 0.11$ &  $4.49 \pm 0.12$ &   $0.82 \pm 0.18$ \\
      & 128 &  $3.14 \pm 0.71$ &   $4.70 \pm 0.72$ &  $0.58 \pm 0.07$ &  $0.08 \pm 0.00$ &  $0.01 \pm 0.00$ &     $4.00 \pm 0.12$ &  $4.36 \pm 0.11$ &   $0.69 \pm 0.16$ \\
      & 256 &  $2.99 \pm 0.85$ &   $4.63 \pm 0.78$ &  $0.52 \pm 0.07$ &  $0.08 \pm 0.00$ &  $0.01 \pm 0.00$ &     $3.95 \pm 0.12$ &  $4.24 \pm 0.11$ &   $0.76 \pm 0.13$ \\
      & 512 &  $\bm{2.99 \pm 0.69}$ &   $\bm{4.51 \pm 0.80}$ &  $\bm{0.51 \pm 0.04}$ &  $\bm{0.07 \pm 0.00}$ &  $0.01 \pm 0.00$ &     $3.96 \pm 0.12$ &  $4.14 \pm 0.09$ &   $\bm{0.71 \pm 0.16}$ \\

\bottomrule
\end{tabular}

        \vspace{10ex}
        \caption{Test \mnll of Bayesian \dnn with 2 hidden layers on regression datasets. \textsc{nf}: number of hidden features}
        \begin{tabular}{ll||rrrrrrrr}
      \toprule
            & {}      & \multicolumn{8}{r}{test mnll}                                                                                                                                                                    \\
            & dataset & boston                        & concrete             & energy               & kin8nm                & naval                 & powerplant           & protein              & yacht                \\
      model & nf      &                               &                      &                      &                       &                       &                      &                      &                      \\
      \midrule
      mcd & 64  &  $5.67 \pm 2.35$ &  $3.19 \pm 0.28$ &  $4.19 \pm 0.15$ &  $-0.78 \pm 0.69$ &  $-2.68 \pm 0.00$ &     -- &  $2.79 \pm 0.01$ &  $2.85 \pm 1.02$ \\
      & 128 &  $6.90 \pm 2.93$ &  $3.20 \pm 0.36$ &  $4.15 \pm 0.15$ &  $-0.87 \pm 0.02$ &  $-1.00 \pm 2.27$ &     $2.74  \pm  0.05$ &  $2.76 \pm 0.02$ &  $2.95 \pm 1.27$ \\
      & 256 &  $6.60 \pm 3.59$ &  $3.31 \pm 0.45$ &  $4.13 \pm 0.15$ &  $-0.70 \pm 0.05$ &  $-2.70 \pm 0.00$ &  $ 2.75  \pm  0.04 $ &  $2.72 \pm 0.01$ &  $3.79 \pm 1.88$ \\
      & 512 &  $7.28 \pm 3.31$ &  $3.45 \pm 0.59$ &  $4.13 \pm 0.17$ &  $-0.76 \pm 0.03$ &  $-2.71 \pm 0.00$ &     $2.77  \pm  0.04$ &  $\bm{2.68 \pm 0.02}$ &  $3.76 \pm 1.65$ \\
mfg & 64  &  $2.83 \pm 0.33$ &  $3.26 \pm 0.08$ &  $4.42 \pm 0.10$ &  $-0.92 \pm 0.02$ &  $-6.24 \pm 0.01$ &      $2.80 \pm 0.03$ &  $2.90 \pm 0.01$ &  $2.85 \pm 0.24$ \\
      & 128 &  $2.99 \pm 0.41$ &  $3.41 \pm 0.05$ &  $4.91 \pm 0.09$ &  $-0.83 \pm 0.02$ &  $-6.23 \pm 0.01$ &      $2.83 \pm 0.03$ &  $2.92 \pm 0.01$ &  $3.38 \pm 0.29$ \\
      & 256 &  $3.33 \pm 0.53$ &  $3.57 \pm 0.07$ &  $5.44 \pm 0.05$ &  $-0.69 \pm 0.01$ &  $-6.22 \pm 0.01$ &      $2.89 \pm 0.02$ &  $2.95 \pm 0.01$ &  $3.65 \pm 0.32$ \\
      & 512 &  $3.69 \pm 0.54$ &  $3.73 \pm 0.05$ &  $5.83 \pm 0.05$ &  $-0.49 \pm 0.01$ &  $-6.19 \pm 0.01$ &      $3.04 \pm 0.03$ &  $2.98 \pm 0.01$ &  $3.86 \pm 0.31$ \\
nng & 64  &  $\bm{2.69 \pm 0.06}$ &  $3.40 \pm 0.15$ &  $\bm{1.95 \pm 0.08}$ &  $-1.14 \pm 0.05$ &  $-5.83 \pm 1.49$ &      $2.80 \pm 0.01$ &  $2.78 \pm 0.01$ &  $2.71 \pm 0.17$ \\
      & 128 &  $2.72 \pm 0.09$ &  $3.56 \pm 0.08$ &  $2.11 \pm 0.12$ &  $-1.19 \pm 0.04$ &  $\bm{-6.52 \pm 0.09}$ &      $2.86 \pm 0.02$ &  $2.95 \pm 0.12$ &  $3.06 \pm 0.27$ \\
      & 256 &  $3.04 \pm 0.22$ &  $3.52 \pm 0.07$ &  $2.64 \pm 0.17$ &  $-1.19 \pm 0.03$ &  $-5.73 \pm 0.21$ &      $2.84 \pm 0.00$ &  $3.02 \pm 0.01$ &  $3.15 \pm 0.13$ \\
      & 512 &  $3.13 \pm 0.14$ &  $3.91 \pm 0.20$ &  $3.07 \pm 0.07$ &  $-0.80 \pm 0.00$ &  $-5.30 \pm 0.05$ &      $3.51 \pm 0.00$ &            -- &  $3.21 \pm 0.14$ \\
whvi & 64  &  $3.68 \pm 1.40$ &  $3.19 \pm 0.34$ &  $2.18 \pm 0.37$ &  $-1.13 \pm 0.02$ &  $-6.25 \pm 0.01$ &      $2.73 \pm 0.03$ &  $2.82 \pm 0.01$ &  $2.56 \pm 1.33$ \\
      & 128 &  $4.33 \pm 1.80$ &  $\bm{3.17 \pm 0.37}$ &  $2.00 \pm 0.60$ &  $-1.19 \pm 0.04$ &  $-6.25 \pm 0.01$ &      $2.71 \pm 0.03$ &  $2.79 \pm 0.01$ &  $1.80 \pm 1.01$ \\
      & 256 &  $4.99 \pm 2.65$ &  $3.35 \pm 0.59$ &  $2.06 \pm 0.72$ &  $\bm{-1.23 \pm 0.04}$ &  $-6.25 \pm 0.01$ &      $\bm{2.70 \pm 0.03}$ &  $2.77 \pm 0.01$ &  $1.53 \pm 0.53$ \\
      & 512 &  $5.41 \pm 2.30$ &  $3.33 \pm 0.56$ &  $2.05 \pm 0.46$ &  $-1.22 \pm 0.04$ &  $-6.25 \pm 0.01$ &      $\bm{2.70 \pm 0.03}$ &  $2.74 \pm 0.01$ &  $\bm{1.37 \pm 0.57}$ \\

      \bottomrule
\end{tabular}

    \end{table}
    \begin{table}
        \sc
        \tiny
        \centering
        \setlength\tabcolsep{4pt}
        \caption{Results of Bayesian \dnn on 6 classification datasets. Note: \textsc{nl}: number of hidden layers, \textsc{nf}: number of hidden features}
        \begin{tabular}{lll||rrrrrr|rrrrrr}
      \toprule
            &    & {}      & \multicolumn{6}{r|}{test error} & \multicolumn{6}{r}{test mnll}                                                                                                                                                                                                                                  \\
            &    & dataset & drive                           & eeg                           & letter               & magic                & miniboo              & mocap                & drive                & eeg                  & letter               & magic                & miniboo              & mocap           \\
      model & nl & nf      &                                 &                               &                      &                      &                      &                      &                      &                      &                      &                      &                      &                 \\
      \midrule
      mcd   & 2  & 64      & $0.19 \pm 0.11$                 & $0.16 \pm 0.01$               & $0.45 \pm 0.05$      & $0.13 \pm 0.02$      & $\bm{0.07 \pm 0.00}$ & $0.02 \pm 0.02$      & $0.52 \pm 0.24$      & $0.36 \pm 0.02$      & $1.27 \pm 0.26$      & $0.37 \pm 0.12$      & $0.18 \pm 0.00$      & $0.11 \pm 0.10$ \\
            &    & 128     & $0.17 \pm 0.07$                 & $0.19 \pm 0.11$               & $0.45 \pm 0.04$      & $0.16 \pm 0.08$      & $0.15 \pm 0.21$      & $0.04 \pm 0.07$      & $0.47 \pm 0.19$      & $0.36 \pm 0.09$      & $1.39 \pm 0.22$      & $0.33 \pm 0.04$      & $0.24 \pm 0.17$      & $0.10 \pm 0.11$ \\
            &    & 256     & $0.16 \pm 0.09$                 & $0.20 \pm 0.15$               & $0.45 \pm 0.06$      & $\bm{0.13 \pm 0.01}$ & $0.07 \pm 0.00$      & $0.16 \pm 0.13$      & $0.50 \pm 0.29$      & $\bm{0.33 \pm 0.08}$ & $1.32 \pm 0.25$      & $0.35 \pm 0.09$      & $\bm{0.17 \pm 0.00}$ & $0.29 \pm 0.21$ \\
            &    & 512     & $0.18 \pm 0.11$                 & $0.18 \pm 0.15$               & $0.44 \pm 0.02$      & $0.18 \pm 0.10$      & $\bm{0.07 \pm 0.00}$ & $0.03 \pm 0.06$      & $0.47 \pm 0.27$      & $0.95 \pm 1.63$      & $1.41 \pm 0.17$      & $0.40 \pm 0.06$      & $0.20 \pm 0.04$      & $0.17 \pm 0.22$ \\
            & 3  & 64      & $0.34 \pm 0.10$                 & $\bm{0.13 \pm 0.01}$          & $0.50 \pm 0.06$      & $0.16 \pm 0.07$      & $0.08 \pm 0.02$      & $0.09 \pm 0.09$      & $0.88 \pm 0.25$      & $0.55 \pm 0.61$      & $1.56 \pm 0.28$      & $0.42 \pm 0.16$      & $0.20 \pm 0.05$      & $0.18 \pm 0.15$ \\
            &    & 128     & $0.32 \pm 0.10$                 & $0.21 \pm 0.14$               & $0.48 \pm 0.09$      & $0.16 \pm 0.07$      & $0.23 \pm 0.28$      & $0.11 \pm 0.19$      & $0.86 \pm 0.28$      & $1.46 \pm 2.78$      & $1.40 \pm 0.34$      & $0.44 \pm 0.13$      & $0.28 \pm 0.18$      & $0.34 \pm 0.28$ \\
            &    & 256     & $0.32 \pm 0.21$                 & $0.23 \pm 0.17$               & $0.43 \pm 0.05$      & $0.14 \pm 0.00$      & $0.23 \pm 0.28$      & $0.28 \pm 0.26$      & $0.87 \pm 0.51$      & $0.40 \pm 0.09$      & $1.34 \pm 0.19$      & $0.62 \pm 0.07$      & $0.31 \pm 0.22$      & $0.61 \pm 0.48$ \\
            &    & 512     & $0.36 \pm 0.09$                 & $0.14 \pm 0.11$               & $0.49 \pm 0.06$      & $0.14 \pm 0.01$      & $0.23 \pm 0.28$      & $0.23 \pm 0.12$      & $0.93 \pm 0.27$      & $0.74 \pm 0.78$      & $1.92 \pm 0.23$      & $1.02 \pm 0.15$      & $0.30 \pm 0.20$      & $0.45 \pm 0.27$ \\
      \hvi  & 2  & 64      & $0.03 \pm 0.01$                 & $0.25 \pm 0.01$               & $0.43 \pm 0.01$      & $\bm{0.13 \pm 0.01}$ & $0.10 \pm 0.00$      & $0.08 \pm 0.01$      & $0.14 \pm 0.04$      & $0.61 \pm 0.28$      & $1.07 \pm 0.02$      & $0.32 \pm 0.02$      & $0.23 \pm 0.01$      & $0.28 \pm 0.02$ \\
            &    & 128     & $0.02 \pm 0.00$                 & $0.21 \pm 0.01$               & $0.41 \pm 0.01$      & $\bm{0.13 \pm 0.01}$ & $0.09 \pm 0.00$      & $0.05 \pm 0.00$      & $0.09 \pm 0.02$      & $0.45 \pm 0.01$      & $1.02 \pm 0.02$      & $0.32 \pm 0.02$      & $0.22 \pm 0.01$      & $0.17 \pm 0.01$ \\
            &    & 256     & $\bm{0.01 \pm 0.00}$            & $0.19 \pm 0.01$               & $0.40 \pm 0.01$      & $\bm{0.13 \pm 0.01}$ & $0.08 \pm 0.00$      & $0.03 \pm 0.00$      & $0.09 \pm 0.03$      & $0.76 \pm 0.92$      & $0.99 \pm 0.01$      & $0.31 \pm 0.02$      & $0.20 \pm 0.00$      & $0.12 \pm 0.01$ \\
            &    & 512     & $\bm{0.01 \pm 0.00}$            & $0.17 \pm 0.01$               & $0.40 \pm 0.01$      & $\bm{0.13 \pm 0.01}$ & $0.08 \pm 0.00$      & $\bm{0.02 \pm 0.00}$ & $0.08 \pm 0.03$      & $0.52 \pm 0.37$      & $0.97 \pm 0.01$      & $\bm{0.31 \pm 0.01}$ & $0.19 \pm 0.01$      & $0.08 \pm 0.01$ \\
            & 3  & 64      & $0.03 \pm 0.00$                 & $0.33 \pm 0.05$               & $0.42 \pm 0.01$      & $\bm{0.13 \pm 0.01}$ & $0.10 \pm 0.00$      & $0.07 \pm 0.01$      & $0.12 \pm 0.02$      & $0.61 \pm 0.05$      & $1.02 \pm 0.02$      & $0.32 \pm 0.01$      & $0.23 \pm 0.01$      & $0.24 \pm 0.02$ \\
            &    & 128     & $0.02 \pm 0.00$                 & $0.38 \pm 0.09$               & $0.41 \pm 0.01$      & $\bm{0.13 \pm 0.01}$ & $0.09 \pm 0.00$      & $0.04 \pm 0.00$      & $0.09 \pm 0.02$      & $0.64 \pm 0.07$      & $0.98 \pm 0.01$      & $0.31 \pm 0.02$      & $0.22 \pm 0.01$      & $0.15 \pm 0.01$ \\
            &    & 256     & $0.05 \pm 0.09$                 & $0.45 \pm 0.01$               & $0.39 \pm 0.01$      & $\bm{0.13 \pm 0.01}$ & $0.08 \pm 0.00$      & $\bm{0.02 \pm 0.00}$ & $0.20 \pm 0.34$      & $0.69 \pm 0.00$      & $0.94 \pm 0.02$      & $0.31 \pm 0.02$      & $0.20 \pm 0.01$      & $0.09 \pm 0.01$ \\
            &    & 512     & $\bm{0.01 \pm 0.00}$            & $0.45 \pm 0.01$               & $\bm{0.38 \pm 0.01}$ & $\bm{0.13 \pm 0.01}$ & $0.08 \pm 0.00$      & $\bm{0.02 \pm 0.00}$ & $\bm{0.05 \pm 0.02}$ & $0.69 \pm 0.00$      & $\bm{0.90 \pm 0.01}$ & $0.32 \pm 0.01$      & $0.19 \pm 0.01$      & $\bm{0.06 \pm 0.01}$ \\
      \bottomrule
\end{tabular}

    \end{table}
\end{landscape}

\end{document}